\documentclass[10pt, journal]{IEEEtran}

\usepackage{caption}
\usepackage[numbers,sort&compress]{natbib}
\usepackage{bm}
\usepackage{amssymb}

\usepackage{amsmath}
\usepackage{graphicx}
\usepackage{subfig}
\usepackage{multirow}
\usepackage{tabularx}
\usepackage{booktabs}

\usepackage[ruled,vlined]{algorithm2e}
\begin{document}

\captionsetup[figure]{labelformat={default},labelsep=period,name={Fig.}}
\captionsetup[table]{justification=centering, labelformat=simple, labelsep=newline, textfont=sc}
\captionsetup{font={small}}

% ----------------------------title------------------------------

\title{Hyper RPCA: Joint Maximum Correntropy Criterion and Laplacian Scale Mixture Modeling On-the-Fly for Moving Object Detection}

\author{Zerui~Shao,
        Yifei~Pu,
        Jiliu~Zhou,~\IEEEmembership{Senior Member,~IEEE,}
        Bihan~Wen,~\IEEEmembership{Member,~IEEE,}
        and~Yi~Zhang,~\IEEEmembership{Member,~IEEE}
        \thanks{Corresponding authors: Bihan~Wen~and~Yi~Zhang.}
        \thanks{Z.~Shao, Y.~Pu, J.~Zhou and Y.~Zhang are with the College of Computer Science, Sichuan University, Chengdu 610065, China (e-mail: zeruishao@outlook.com; puyifei@scu.edu.cn; zhoujl@scu.edu.cn; yzhang@scu.edu.cn).}
        \thanks{B.~Wen is with the School of Electrical and Electronic Engineering, Nanyang Technological University, Singapore 639798 (e-mail: bihan.wen@ntu.edu.sg).}
}

% \markboth{IEEE TRANSACTIONS ON COMPUTATIONAL IMAGING}%
% {SHAO \MakeLowercase{\textit{et al.}}: Hyper RPCA: Joint Maximum Correntropy Criterion and Laplacian Scale Mixture Modeling On-the-Fly for Moving Object Detection}

\maketitle

% ----------------------------abstract------------------------------

\begin{abstract}
Moving object detection is critical for automated video analysis in many vision-related tasks, such as surveillance tracking, video compression coding, etc. Robust Principal Component Analysis (RPCA), as one of the most popular moving object modelling methods, aims to separate the temporally-varying (i.e., moving) foreground objects from the static background in video, assuming the background frames to be low-rank while the foreground to be spatially sparse. Classic RPCA imposes sparsity of the foreground component using $\ell_1$-norm, and minimizes the modeling error via $\ell_2$-norm. We show that such assumptions can be too restrictive in practice, which limits the effectiveness of the classic RPCA, especially when processing videos with dynamic background, camera jitter, camouflaged moving object, etc. In this paper, we propose a novel RPCA-based model, called Hyper RPCA, to detect moving objects on the fly. Different from classic RPCA, the proposed Hyper RPCA jointly applies the maximum correntropy criterion (MCC) for the modeling error, and Laplacian scale mixture (LSM) model for foreground objects. Extensive experiments have been conducted, and the results demonstrate that the proposed Hyper RPCA has competitive performance for foreground detection to the state-of-the-art algorithms on several well-known benchmark datasets.
\end{abstract}


% ----------------------------Introduction------------------------------

\section{Introduction}
\IEEEPARstart{M}{oving} object detection (MOD) of surveillance video frames is critical for many computer vision applications, such as video compression coding, object behavior extraction and surveillance object tracking \cite{paul2010video,chen2012surveillance,jodoin2012behavior,cullen2012detection,yilmaz2006object}. In the past decades, various MOD methods \cite{oreifej2012simultaneous,xu2013gosus,gao2014block,xin2015background,gemignani2016robust,javed2016spatiotemporal,bouwmans2017decomposition,javed2017background} have been proposed. Early strategies proposed to directly distinguish background pixels from foreground through simple statistical measures, such as median, mean and histogram model \cite{zheng2006extracting}. Later, more elaborate methods proposed to classify the pixels by learning background and foreground models, such as the Gaussian mixture models (GMM) \cite{zivkovic2004improved} and local binary pattern (LBP) models \cite{heikkila2006texture}. However, these methods failed to exploit critical structures, such as temporal similarity between frames, or sparsity of foreground objects. When processing complex video data involving camera jitter, dynamic background and illumination, etc., the performance would degrade significantly. Recently, based on the assumption that the background is low-rank and the foreground is sparse, the RPCA-based approaches \cite{candes2011robust,zhou2012moving,feng2013online,shakeri2016corola,vaswani2018robust,bouwmans2018applications} have attracted much attention for background subtraction. By exploiting the low-rank property of the background and the sparse prior of the foreground, the RPCA-based methods have achieved a remarkable success in MOD tasks.

Despite the success of the classic RPCA-based methods \cite{candes2011robust,feng2013online}, they suffer from two major limitations for MOD in practice: (1) The sparsity assumption, which is imposed by $\ell_1$-norm penalty, is sometimes too restrictive for large and complicated foreground in practice. (2) It is unclear whether the $\ell_2$-norm is the optimal penalty for the RPCA modeling error, which highly depends on the model accuracy and video data distribution. Various works have been proposed to tackle the limitation (1) towards more effective foreground object modeling \cite{javed2018moving,zhou2012moving,shi2018robust,yong2017robust,ebadi2017foreground}. For example, \cite{zhou2012moving} applied Markov Random Field (MRF) to constrain the sparse foreground parts, and $\ell_0$-norm is utilized to regularize the sparse component. \cite{ebadi2017foreground} was inspired by the concept of the group sparsity structure, and exploited the tree-structured property of the moving objects. Instead of using fixed foreground distribution, \cite{yong2017robust} proposed to model the foreground pixel with separate mixture of Gaussian (MoG) distribution. In \cite{javed2018moving}, graph Laplacian was imposed in both spatial and temporal domains to explore the spatiotemporal correlation in sparse component. The Gaussian scale mixture (GSM) model was utilized in \cite{shi2018robust} to model each foreground pixel, which significantly improved the estimation accuracy by jointly estimating the variances of the known and unknown sparse coefficients. On the contrary, few works addressed the limitation (2) on modeling error. Very recently, \cite{guo2017godec+} applied the maximum correntropy criterion (MCC) \cite{he2011robust} for modeling error, to obtain higher-quality background extraction. However, \cite{guo2017godec+} is a batch algorithm, thus scales poorly for real-time video MOD.

In this paper, we propose a novel online MOD scheme via background subtraction, dubbed Hyper RPCA. The proposed scheme simultaneously tackles the two aforementioned limitations of the classic RPCA, by jointly applying the Laplacian scale mixture (LSM) and MCC models, to effectively model the complicated foreground moving objects, and approximation error, respectively. Furthermore, the proposed Hyper RPCA can be learned and applied on the fly (i.e., process video streaming sequentially) with high scalability and low latency, which is more efficient for processing video streams from surveillance cameras in practice. Our contributions in this paper are summarized as follows.

\begin{itemize}
	\item We propose a novel Hyper RPCA formulation, which combines the LSM model and MCC respectively for complicated foreground moving objects and accurate error modeling. Compared with classic RPCA, the proposed formulation can effectively model foreground pixels and is capable of adapting a wide range of modeling error.
	
	\item A highly-efficient online algorithm to solve Hyper RPCA for MOD is derived. The proposed online algorithm avoids high computational complexity and can be used for real-time MOD applications.
	
	\item  Extensive experiments over several datasets of challenging scenarios are conducted. The results demonstrate that the proposed Hyper RPCA outperforms the state-of-the-art MOD methods over several benchmark datasets.
\end{itemize}

The remainder of the paper is organized as follows. Section \ref{sec.related works} reviews the related works to the Hyper RPCA. Section \ref{sec.Hyper RPCA} presents the proposed Hyper RPCA learning formulation. Section \ref{sec.algorithm} describes the highly-efficient Hyper RPCA algorithm for MOD. Section \ref{sec.experiments} presents the experimental results on several challenging MOD cases. Section \ref{sec.conclusion} concludes this paper.

% ----------------------------Related Works------------------------------

\section{Related Works}\label{sec.related works}
\subsection{Online RPCA}
While the classic RPCA process batch data, recent works proposed online RPCA schemes for more efficient and scalable MOD \cite{zhan2016online,javed2018moving,shi2018robust,feng2013online,lois2015online}. The idea is to decompose the nuclear norm of batch RPCA \cite{candes2011robust} to an explicit product of two low-rank matrices, then the objective function can be solved by stochastic optimization algorithm \cite{feng2013online}, i.e., the coefficient and sparse component of each frame are updated by the previous basis, and then the basis is updated alternately. Comparing to the batch RPCA, the online RPCA has lower latency, thus is more suitable for video MOD.

\subsection{Laplacian Scale Mixture (LSM) Model}
LSM model has demonstrated its potential for sparse signal modeling in recent works \cite{garrigues2010group,huang2017mixed,dong2018robust}: In \cite{garrigues2010group}, LSM model has been used to model the dependencies among sparse coding coefficients for image coding and compressive sensing recovery. The basic idea of LSM model is to model each sparse pixel as a product of a random Laplacian variable and a positive hidden multiplier, and impose a hyperprior (i.e., the Jeffrey prior \cite{gep1973bayesian} used in this paper) over the positive hidden multiplier. These models allow to jointly estimate both sparse pixels and hidden multipliers from the observed data under the MAP estimation framework via alternative optimization. In \cite{huang2017mixed}, the LSM model was used for mixed noise removal, where the impulse noise is modeled with LSM distributions. In \cite{dong2018robust}, the tensor coefficients were modeled with LSM for multi-frame image and video denoising. By jointly estimating the variances and recovering the values of coefficients, LSM-based methods significantly improved the modeling accuracy from plain sparse coding with more flexibility and robustness.

\subsection{Maximum Correntropy Criterion (MCC)}
Correntropy \cite{liu2007correntropy} is known for more effectively modeling error or noise beyond Gaussian distribution in practice \cite{guo2017godec+}. It models a nonlinear similarity between two random variables $X$ and $Y$ as $V_{\sigma}(X,Y)=E[k_{\sigma}(X,Y)]=\int k_{\sigma}(x,y)dF_{XY}(x,y)$, where $E[\cdot]$ is the expectation operator, $F_{XY}(x,y)$ is the joint distribution function of $(X,Y)$, and $k_{\sigma}(\cdot)$ is the kernel function, we select Gaussian kernel as the kernel function, so $k_{\sigma}(e)=g_{\sigma}(e)=\exp(-e^{2}/(2\sigma^{2}))$, where $e=x-y$, $\sigma$ is the kernel width and controls the radial range of the Gaussian kernel function, and $\exp(\cdot)$ denotes the exponential operation on each element of the input parameters. As the joint distribution is mostly unknown in practice, the correntropy of $(X,Y)$ can be approximated by $\hat{V}_{n,\sigma}(X,Y)=(1/n)\sum\nolimits_{i=1}^{n}\left[ g_{\sigma}(x_{i}-y_{i}) \right]$, given a finite number of samples $\{ (x_{i},y_{i}) \}_{i=1}^{n}$, and accordingly the MCC is equivalent to minimizing $(1/n)\sum\nolimits_{i=1}^{n}\sigma^{2}(1-g_{\sigma}(x_{i}-y_{i}))$ \cite{he2019robust}. Recent works applied MCC for a wide range of important applications, including face recognition, matrix completion, and low-rank matrix decomposition \cite{he2011robust,guo2017godec+,he2019robust,he2010maximum,du2017robust,lu2013correntropy}, which is shown to be highly effective.

% ----------------------------Hyper RPCA------------------------------

\section{Hyper RPCA Learning Formulation}\label{sec.Hyper RPCA}
We first demonstrate the major limitations of classic RPCA and then introduce to the LSM and MCC, based on which we propose the Hyper RPCA formulation.

\begin{figure}[htpb]
	\centering
	\includegraphics[width=.95\linewidth]{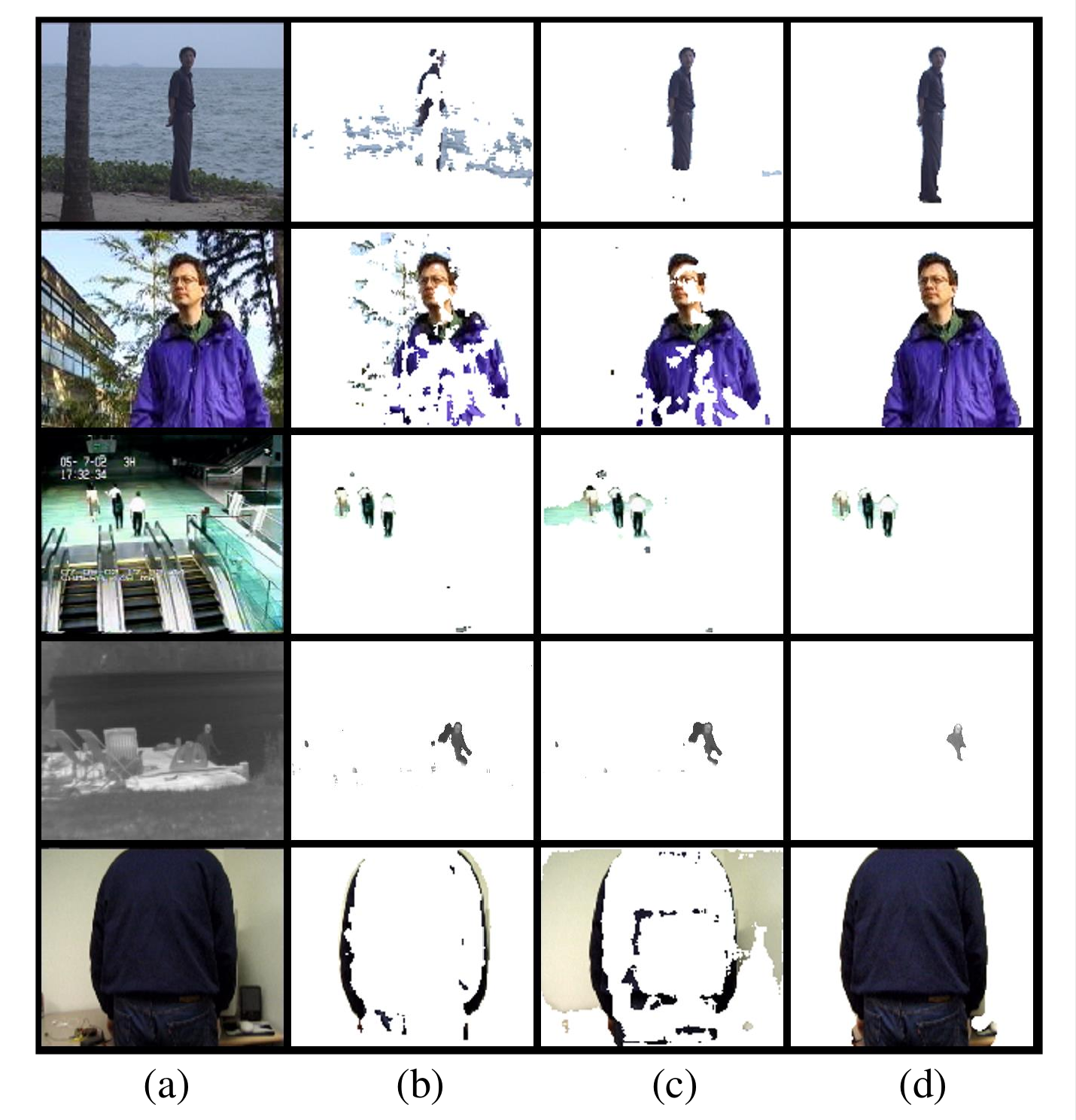}
	\caption{Visual results of different methods for foreground detection. (a) Input images, (b) PCP \cite{candes2011robust}, (c) ORPCA \cite{feng2013online} and (d) the proposed Hyper RPCA method. From top to bottom: the 1541th frame of WaterSurface sequence from I2R dataset \cite{li2004statistical}, the 1248th frame of WavingTrees sequence from Wallflower dataset \cite{toyama1999wallflower}, the 301th frame of Escalator sequence from I2R dataset, the 1220th frame of lakeSide sequence from CDnet dataset \cite{wang2014cdnet}, the 253th frame of Camouflage sequence from Wallflower dataset.
	\label{fig.1}
	}
\end{figure}

\subsection{Preliminary}
The classic online RPCA method \cite{feng2013online} assumes that background is low-rank and foreground is sparse simultaneously. It utilizes $\ell_{2}$-norm and $\ell_{1}$-norm for modeling approximation error, and the sparsity of the foreground, respectively. The formulation of the classic online RPCA is the following

\begin{equation}\label{eq.1}
	\begin{split}
		(\bm{U},\bm{V},\bm{S})=&\mathop{\arg\min}_{\bm{U},\bm{V},\bm{S}}
		\|\bm{D}-\bm{UV}^\top-\bm{S}\|_{F}^{2} + \\
		&\eta\{ \|\bm{U}\|_F^2 + \|\bm{V}\|_F^2 \} + 
		\lambda\|\bm{S}\|_1\,\text{,}
	\end{split}
\end{equation}

\noindent where $\bm{D}=[\bm{d}_1,\bm{d}_2,\cdots,\bm{d}_T]\in\mathbb{R}^{p \times T}$ is the data matrix of $T$ frames, i.e., $\bm{d}_t\in\mathbb{R}^p$ denotes the $t$-th frame, and $p=m \times n$ denotes the frame size. The foreground and background components are denoted as $\bm{L}=\bm{UV}^\top=[\bm{l}_1,\bm{l}_2,\cdots,\bm{l}_T]\in\mathbb{R}^{p \times T}$ and $\bm{S}=[\bm{s}_1,\bm{s}_2,\cdots,\bm{s}_T]\in\mathbb{R}^{p \times T}$, respectively. As the background is assumed low-rank, $\bm{U}\in\mathbb{R}^{p \times r}$ and $\bm{V}\in\mathbb{R}^{T \times r}$are used to represent the background basis and coefficients, respectively, with $r \ll p,T$, thus, $\bm{L=UV}^\top$ is the low-rank approximation of the background. In practice, the RPCA assumption sometimes becomes too restrictive:

\subsubsection{}
The foreground may not be sufficiently sparse in the spatial domain. Fig. \ref{fig.1} shows the foreground extraction results of some example video frames from three challenging datasets \cite{wang2014cdnet,li2004statistical,toyama1999wallflower}, using PCP \cite{candes2011robust}, ORPCA \cite{feng2013online} and the proposed Hyper RPCA method. Comparing to the proposed Hyper RPCA using more complicated LSM, the sparse modeling based on $\ell_{1}$-norm fails to capture the complete foreground objects effectively, e.g., dynamic background (rows 1 and 2), small moving objects (rows 3 and 4) or large-area foreground objects relative to the background (row 5) in which the moving objects is not sparse spatially.

\subsubsection{}
\begin{figure*}[htbp]
	\centering
	\subfloat[]{
		\begin{minipage}[]{0.3\linewidth}
			\centering
			\includegraphics[width=5cm]{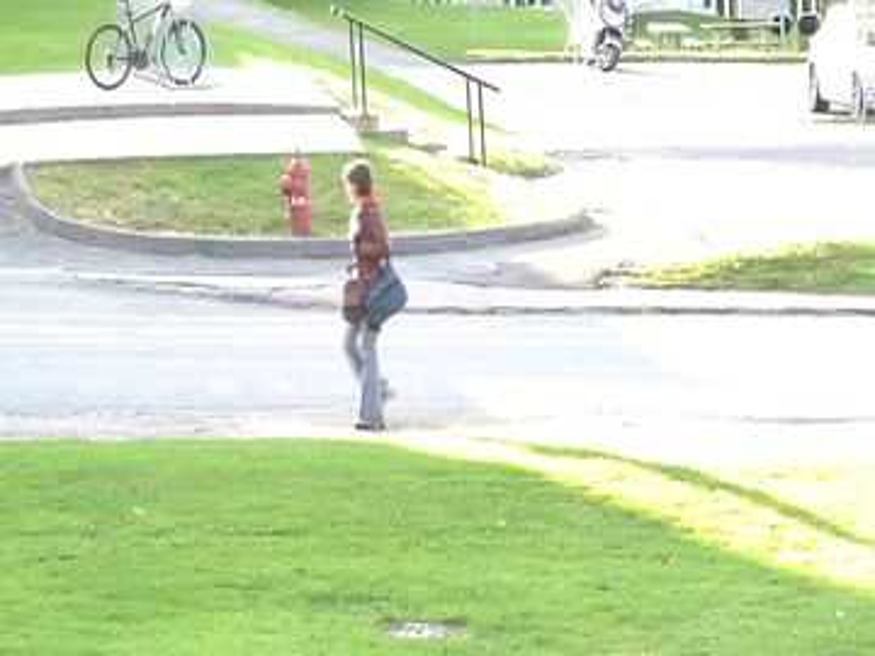}
			%\caption{fig1}
		\end{minipage}%
	}%
	\subfloat[]{
		\begin{minipage}[]{0.3\linewidth}
			\centering
			\includegraphics[width=5cm]{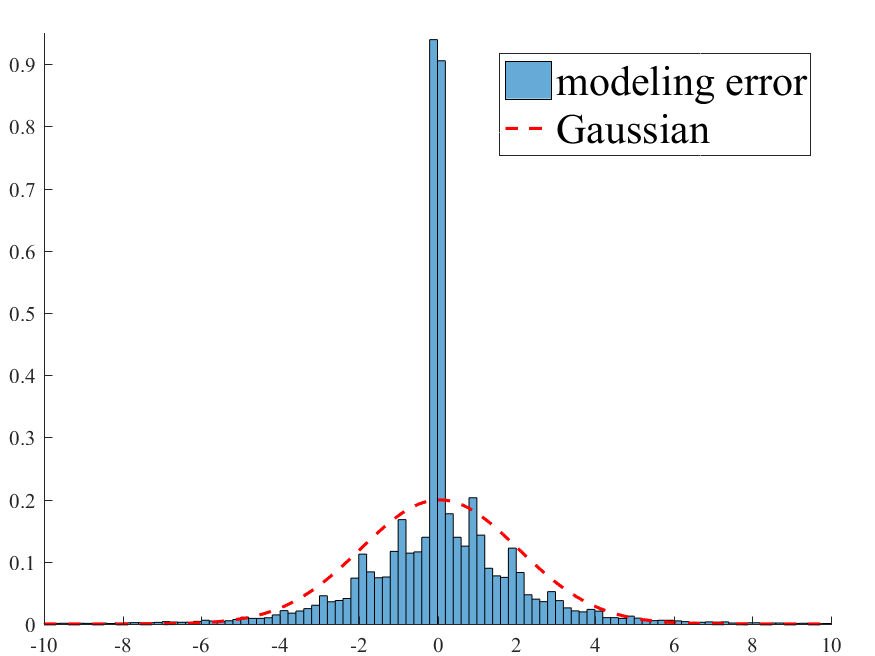}
			%\caption{fig1}
		\end{minipage}%
	}%
	\subfloat[]{
		\begin{minipage}[]{0.3\linewidth}
			\centering
			\includegraphics[width=5cm]{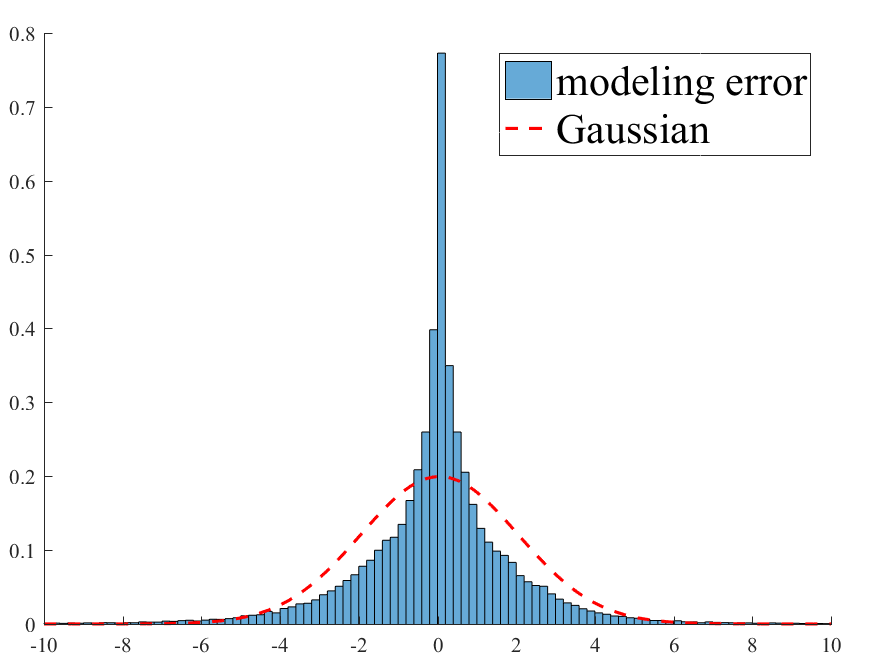}
			%\caption{fig1}
		\end{minipage}%
	}%
	\caption{The distributions of modeling error on Pedestrians test sequence. (a) The 960th frame of Pedestrians sequence from ``Baseline'' category in CDnet dataset \cite{wang2014cdnet}. The distribution results generated by (b) PCP \cite{candes2011robust} and (c) ORPCA \cite{feng2013online}.}
	\label{fig.2}
\end{figure*}

The real video data is often corrupted by complicated or hybrid types of noise, thus the modeling error is deviated from Gaussian distribution (which is usually modeled using $\ell_{2}$-norm). Fig. \ref{fig.2} presents an analysis of the modeling error distribution of an example video \cite{wang2014cdnet} (a) using the classic RPCA-based method \cite{candes2011robust,feng2013online}. Fig. \ref{fig.2} (b) and (c) plot their empirical distribution of modeling error, respectively, which both deviate from Gaussian distribution.

\subsection{Hyper RPCA for moving object detection}
We propose the novel Hyper RPCA, to tackle the aforementioned limitations of the classic online RPCA-based method, by jointly applying MCC and LSM models. The batch learning formulation of Hyper RPCA is the following

% \sqrt{\bm{W}}

\begin{equation}\label{eq.2}
	\begin{split}
		\mathop{\arg\min}_{\bm{W},\bm{U},\bm{V},\bm{B},\bm{A}} 
		&\|\bm{W}^{1/2}\odot(\bm{D}-\bm{UV}^\top-\bm{B}\odot\bm{A})\|_F^2 \\ 
		& + \eta\sigma_{w}^2 \{ \|\bm{U}\|_F^2 + \|\bm{V}\|_F^2 \} + 
		2\sigma_{w}^2\sum\nolimits_{t}\sum\nolimits_{i}\vert\alpha_{i,t}\vert \\ 
		& + 4\sigma_{w}^2\sum\nolimits_{t}\sum\nolimits_{i}\log(b_{i,t}+\varepsilon)\,\text{,}
	\end{split}
\end{equation}

\noindent where $\bm{B}=[\bm{b}_1,\bm{b}_2,\cdots,\bm{b}_T]\in\mathbb{R}^{p \times T}$ denotes the matrix of positive hidden multipliers $b_{i,t}$, similarly, $\bm{A}=[\bm{\alpha}_1,\bm{\alpha}_2,\cdots,\bm{\alpha}_T]\in\mathbb{R}^{p \times T}$ is the matrix representation of the Laplacian variables $\alpha_{i,t}$, $\bm{W}$ is the weight matrix, $\sigma_w^2$ is the variance of the modeling error, $\varepsilon$ is a small constant for numerical stability, and $\odot$ denotes element-wise multiplication of two matrices. In addition, the online solution for the proposed method Eq. \ref{eq.2} can be reformulated as

\begin{equation}\label{eq.3}
	\begin{split}
		\mathop{\arg\min}_{\bm{W},\bm{U},\bm{V},\bm{B},\bm{A}} 
		&\sum\nolimits_{i=1}^{T} \{
		\|\bm{w}_t\circ(\bm{d}_t-\bm{Uv}_t-\bm{b}_t\circ\bm{\alpha}_t)\|_2^2 \\ 
		& + \eta\sigma_{w}^2\|\bm{v}_t\|_2^2 
		+ 2\sigma_{w}^2\sum\nolimits_{i}\vert\alpha_{i,t}\vert \\ 
		& + 4\sigma_{w}^2\sum\nolimits_{i}\log(b_{i,t}+\varepsilon) \} 
		+ \eta\sigma_{w}^2\|\bm{U}\|_F^2\,\text{,}
	\end{split}
\end{equation}

\noindent where $\bm{b}_t=[b_{1,t},b_{2,t},\cdots,b_{p,t}]^\top\in\mathbb{R}^p$, $\bm{\alpha}_t=[\alpha_{1,t},\alpha_{2,t},\cdots,\alpha_{p,t}]^\top\in\mathbb{R}^p$, $\bm{v}_t$ and $\bm{w}_t$ are the $t$-th column of $\bm{B}$, $\bm{A}$, $\bm{V}^\top$ and $\bm{W}^{1/2}$, respectively, and $\circ$ denotes element-wise multiplication of two vectors.

\subsection{LSM for foreground modeling}
In realistic scenarios, the prior distribution of foreground $P(\bm{S})$ is unknown and it is difficult to estimate. In LSM modeling, each foreground pixel $s_{i,t}$ is expressed by $s_{i,t}=b_{i,t}\cdot\alpha_{i,t}$, where $\alpha_{i,t}$ is a random Laplacian variable, and $b_{i,t}$ is a positive hidden multiplier. $s_{i,t}$ denotes the $i$-th pixel of $\bm{s}_t$, and is modeled with a zero-mean Laplacian distribution with standard deviation $b_{i,t}$, i.e., $P(s_{i,t} \vert b_{i,t})=(1/2b_{i,t})\exp(-\vert s_{i,t} \vert / b_{i,t})$. A hyper prior $P(b_{i,t})$ is further used to model $b_{i,t}$. Then, the LSM model of $s_{i,t}$ can be expressed as $P(s_{i,t})=\int_{0}^{\infty}P(s_{i,t} \vert b_{i,t})P(b_{i,t})db_{i,t}$, which cannot be expressed in an analytical form in general. Thus, the estimation of $\bm{L}$ and $\bm{S}$ from $\bm{D}$ are considered in the maximum a posterior (MAP) estimator as

\begin{equation}\label{eq.4}
	\begin{split}
		(\bm{L},\bm{S},\bm{B})=\mathop{\arg\max} 
		&\log P(\bm{D} \vert \bm{L},\bm{S}) + \log P(\bm{L}) \\
		& + \log P(\bm{S} \vert \bm{B}) + \log P(\bm{B})\,\text{,}
	\end{split}
\end{equation}

\noindent where $P(\bm{D} \vert \bm{L},\bm{S})$ is the Gaussian likelihood term with mean of zero and variance of $\sigma_w^2$ and $\bm{B}$ denotes the matrix of positive hidden multipliers. In this paper, we use the noninformative Jeffrey's prior $P(b_{i,})=1/b_{i,t}$ to model hidden variable $b_{i,t}$ \cite{gep1973bayesian}, and the prior of $\bm{L}$ is modeled as $P(L)\propto\exp(-\eta \|\bm{L}\|_*)$. By assuming $b_{i,t}$ and $s_{i,t}$ are independent, and $\bm{S}$ is i.i.d., Eq. \ref{eq.4} can be rewritten as

\begin{equation}\label{eq.5}
	\begin{split}
		(\bm{L},\bm{B},\bm{A})=&\mathop{\arg\min}_{\bm{L},\bm{B},\bm{A}} 
		\|\bm{D}-\bm{L}-\bm{B}\odot\bm{A}\|_F^2 \\ 
		& + 2\eta\sigma_w^2\|\bm{L}\|_* 
		+ 2\sigma_w^2\sum\nolimits_{t}\sum\nolimits_{i}\vert \alpha_{i,t} \vert \\ 
		& + 4\sigma_w^2\sum\nolimits_{t}\sum\nolimits_{i}\log(b_{i,t}+\varepsilon)
		\,\text{,}
	\end{split}
\end{equation}

\noindent where $\bm{S}=\bm{B}\odot\bm{A}$ denotes sparse component, and $\bm{A}$ is the matrix representation of the Laplacian variables.

\subsection{MCC for error modeling}
To improve the performance of classic online RPCA-based background subtraction method to deal with non-Gaussian modeling errors, we model this part with correntropy instead of $\ell_2$-norm and Eq. \ref{eq.1} can be rewritten as

\begin{equation}\label{eq.6}
	\begin{split}
		(\bm{U},\bm{V},\bm{S})=&\mathop{\arg\min}_{\bm{U},\bm{V},\bm{S}} 
		\sigma^2[1-g_{\sigma}(\bm{D-L-S})] \\ 
		& + \eta\{ \|\bm{U}\|_F^2 + \|\bm{V}\|_F^2 \} +  \lambda\|\bm{S}\|_1\,\text{,}
	\end{split}
\end{equation}

\noindent where $g_{\sigma}(\cdot)$ denotes the Gaussian kernel operation on each element of the input parameters. Based on the Half-Quadratic (HQ) optimization theory \cite{nikolova2005analysis}, Eq. \ref{eq.6} becomes a weighted matrix factorization problem that can be solved by the strategy used in \cite{he2019robust}. Then, Eq. \ref{eq.6} can be rewritten as

\begin{equation}\label{eq.7}
	\begin{split}
		(\bm{W},\bm{U},\bm{V},\bm{S})=&\mathop{\arg\min}_{\bm{W},\bm{U},\bm{V},\bm{S}}
		\|\bm{W}^{1/2}\odot(\bm{D-L-S})\|_F^2 + \\ 
		& \eta\{ \|\bm{U}\|_F^2 + \|\bm{V}\|_F^2 \} + \lambda\|\bm{S}\|_1\,\text{,}
	\end{split}
\end{equation}

\noindent where $\bm{W}$ is the weight matrix, and its values can be obtained by the modeling error \cite{he2019robust}.

% ----------------------------Algorithm------------------------------

\section{Algorithm}\label{sec.algorithm}
We propose an efficient algorithm solving Eq. \ref{eq.3} by an alternating minimization, the proposed algorithm calculate frame by frame, detect moving objects and gradually ameliorate the background based on the real-time video variations. We describe each sub-problem as follows.
\subsection{Solving the W-Subproblem}
Based on the optimization theory in \cite{he2019robust,nikolova2005analysis}, when the background $\bm{l}_t=\bm{UV}_t$ and $\bm{s}_t=\bm{b}_t \circ \bm{\alpha}_t$ are fixed, for each frame, the optimal solutions of $\bm{w}_t$ can be obtained as

\begin{equation}\label{eq.8}
	\bm{w}_t=sqrt(g_\sigma(\bm{d}_t-\bm{l}_t-\bm{s}_t))\,\text{,}
\end{equation}

\noindent where $sqrt(\cdot)$ denotes the square root operation on each element of input parameters. For a detailed derivation process, please see \cite{he2019robust} for optimizing $\bm{w}_t$.

\subsection{Solving the V-Subproblem}
For fixed $\bm{W}$, $\bm{S}$ and $\bm{U}$, the $\bm{V}$-subproblem can be formulated as

\begin{equation}\label{eq.9}
	\begin{split}
		\bm{V}=\mathop{\arg\min}_{\bm{V}} \sum\nolimits_{t=1}^{T} 
		& \{ \|\bm{w}_t\circ(\bm{d}_t-\bm{Uv}_t-\bm{s}_t)\|_2^2 \\ 
		& + \eta\sigma_w^2\|\bm{v}_t\|_2^2 \}\,\text{,}
	\end{split}
\end{equation}

\noindent For each frame, $\bm{v}_t$ can be obtained by

\begin{equation}\label{eq.10}
	\bm{v}_t=\mathop{\arg\min}_{\bm{v}_t} 
	\|\bm{w}_t\circ(\bm{d}_t-\bm{Uv}_t-\bm{s}_t)\|_2^2 
	+ \eta\sigma_w^2\|\bm{v}_t\|_2^2\,\text{,}
\end{equation}

\noindent which has the closed-form solution as

\begin{equation}\label{eq.11}
	\bm{v}_t=[\bm{U}^\top diag(\bm{w}_t)^2 \bm{U} + \eta\sigma_w^2\bm{I}]^{-1} 
	\bm{U}^\top diag(\bm{w}_t)^2 (\bm{d}_t-\bm{s}_t)\,\text{,}
\end{equation}

\noindent where $diag(\cdot)$ denotes the diagonalization operation for vector, and $\bm{I}$ is an $r \times r$ identity matrix.

\subsection{Solving the S-Subproblem}
For fixed $\bm{W}$ and $\bm{L}$, $\bm{S}$ can be obtained by solving the following optimization problem

\begin{equation}\label{eq.12}
	\begin{split}
		(\bm{B},\bm{A})=&\mathop{\arg\min}_{\bm{B},\bm{A}} \sum\nolimits_{t=1}^{T} 
		\{ \|\bm{w}_t\circ(\bm{d}_t-\bm{Uv}_t-\bm{s}_t)\|_2^2 \\ 
		& + 2\sigma_w^2\sum\nolimits_{i}\vert \alpha_{i,t} \vert 
		+ 4\sigma_w^2\sum\nolimits_{i}\log(b_{i,t}+\varepsilon) \}
		\,\text{,}
	\end{split}
\end{equation}

\subsubsection{Solving for $b_t$}
For each frame, while $\bm{\alpha}_t$ is fixed, $\bm{b}_t$ can be obtained as

\begin{equation}\label{eq.13}
	\begin{split}
		\bm{b}_t&=\mathop{\arg\min}_{\bm{b}_t} 
		\|\bm{w}_t\circ(\bm{d}_t-\bm{l}_t-\bm{b}_t\circ\bm{\alpha}_t)\|_2^2 \\ 
		& \quad + 4\sigma_w^2 \sum\nolimits_{i}\log(b_{i,t}+\varepsilon) \\ 
		&=\mathop{\arg\min}_{\bm{b}_t} \sum\nolimits_{i} \{ 
		(\sqrt{w_{i,t}}(d_{i,t}-l_{i,t}-b_{i,t}\alpha_{i,t}))^2 \\ 
		& \quad + 4\sigma_w^2\log(b_{i,t}+\varepsilon) \}\,\text{,}
	\end{split}
\end{equation}

\noindent Moreover, each $b_{i,t}$ can be solved independently by

\begin{equation}\label{eq.14}
	\begin{split}
		b_{i,t}=&\mathop{\arg\min}_{b_{i,t}} 
		(\sqrt{w_{i,t}}(d_{i,t}-l_{i,t}-b_{i,t}\alpha_{i,t}))^2 \\ 
		& + 4\sigma_w^2\log(b_{i,t}+\varepsilon)\,\text{,}
	\end{split}
\end{equation}

\noindent Though Eq. \ref{eq.14} is nonconvex, the closed-form solution can be obtained by taking $\mathrm{d}f(b_{i,t}) / \mathrm{d}b_{i,t}=0$, where $f(b_{i,t})$ is the right side of Eq. \ref{eq.14}. The solution of Eq. \ref{eq.14} is given by

\begin{equation}\label{eq.15}
	b_{i,t}=
	\begin{cases}
	0&,\text{if $(2a\varepsilon+h)^2/(16a^2)-
		(h\varepsilon+q)/(2a)<0$} \\ 
	T_{i,t}&,\text{otherwise}\,\text{,}
	\end{cases}
\end{equation}

\noindent where $a=(\sqrt{w_{i,t}}\alpha_{i,t})^2$, $h=-2w_{i,t}(d_{i,t}-l_{i,t})\alpha_{i,t}$, $q=4\sigma_w^2$ and $T_{i,t}=\min\{f(0),f(b^{\ast})\}$, where $b^{\ast}$ is the stationary point of $f(b_{i,t})$, which is defined as

\begin{equation}\label{eq.16}
	b^{\ast}=-\frac{2a\varepsilon+h}{4a}\pm\
	\sqrt{\frac{(2a\varepsilon+h)^2}{16a^2}-\frac{h\varepsilon+q}{2a}}
	\,\text{,}
\end{equation}

\subsubsection{Solving for $\alpha_t$}
For fixed $\bm{b}_t$, $\bm{\alpha}_t$ can be solved by minimizing

\begin{equation}\label{eq.17}
	\begin{split}
		\bm{\alpha}_t&=\mathop{\arg\min}_{\bm{\alpha}_t} 
		\|\bm{w}_t\circ(\bm{d}_t-\bm{l}_t-\bm{b}_t\circ\bm{\alpha}_t)\|_2^2 \\ 
		& \quad + 2\sigma_w^2 \sum\nolimits_{i}\vert \alpha_{i,t} \vert \\ 
		&=\mathop{\arg\min}_{\bm{\alpha}_t} \sum\nolimits_{i} \{ 
		(\sqrt{w_{i,t}}(d_{i,t}-l_{i,t}-b_{i,t}\alpha_{i,t}))^2 \\ 
		& \quad + 2\sigma_w^2\vert \alpha_{i,t} \vert \}
		\,\text{,}
	\end{split}
\end{equation}

\noindent Similarly, each $\alpha_{i,t}$ can be solved independently by

\begin{equation}\label{eq.18}
	\alpha_{i,t}=\mathop{\arg\min}_{\alpha_{i,t}} 
	(\sqrt{w_{i,t}}(d_{i,t}-l_{i,t}-b_{i,t}\alpha_{i,t}))^2 
	+ 2\sigma_w^2\vert \alpha_{i,t} \vert
	\,\text{,}
\end{equation}

\noindent which admits a closed-form solution

\begin{equation}\label{eq.19}
	\alpha_{i,t}=S_{\tau_{i,t}}((d_{i,t}-l_{i,t})/(b_{i,t}+\varepsilon))
	\,\text{,}
\end{equation}

\noindent where $S_{\tau_{i,t}}(\cdot)$ is the soft-thresholding operator with threshold $\tau_{i,t}=2\sigma_w^2/(\sqrt{w_{i,t}}b_{i,t}+\varepsilon)^2$. Finally, $\bm{s}_t$ can be computed by $\bm{s}_t=\bm{b}_t\circ\bm{\alpha}_t$ once $\bm{b}_t$ and $\bm{\alpha}_t$ are obtained.

\begin{algorithm}[t]
	\caption{The Hyper RPCA algorithm for MOD}
	\label{alg:1}
	\KwIn{$\bm{D}$,$r$,$\sigma$,$\eta$,$\sigma_w^2$.}
	Initialization: $\bm{U}^{(0)}$ using ($\ref{eq.23}$),$\bm{L}$,$\bm{S}$. \\ 
	
	\For{$t = 1 : T$}{
		%\{Access each sample.\} \\ 
		\While{not converged}{
			Compute $\bm{w}_t$ using (\ref{eq.8}). \\ 
			Compute $\bm{v}_t$ using (\ref{eq.11}). \\ 
			Compute $\bm{b}_t$ using (\ref{eq.15}). \\ 
			Compute $\bm{\alpha}_t$ using (\ref{eq.19}). \\ 
			Compute $\bm{s}_t=\bm{b}_t\circ\bm{\alpha}_t$,
			$\bm{S}(:,\bm{t})\leftarrow\bm{s}_t$. \\ 
		}
		Update $\bm{U}$ using (\ref{eq.22}). \\ 
		Compute $\bm{l}_t=\bm{Uv}_t$,$\bm{L}(:,\bm{t})\leftarrow\bm{l}_t$. \\ 
	}
	\KwOut{$\bm{L}$,$\bm{S}$.}
\end{algorithm}

\subsection{Solving the U-Subproblem}
Similar to \cite{feng2013online}, we use the online learning method to update $\bm{U}$. After estimating $\bm{w}_t$, $\bm{s}_t$ and $\bm{v}_t$, the $\bm{U}$ of $t$-th frame can be updated as

\begin{equation}\label{eq.20}
	\begin{split}
		\bm{U}^{(t)}&\triangleq\mathop{\arg\min}_{\bm{U}} 
		\frac{1}{t}\sum\nolimits_{i=1}^t\frac{1}{2}
		\|\bm{w}_i\circ(\bm{d}_i-\bm{Uv}_i-\bm{s}_i)\|_2^2 \\ 
		& \quad + \frac{\eta\sigma_w^2}{2t}\|\bm{U}\|_F^2 \\ 
		&\triangleq\mathop{\arg\min}_{\bm{U}}
		\frac{1}{2}Tr[\bm{U}^\top(\bm{C}_t+\eta\sigma_w^2\bm{I})\bm{U}] 
		- Tr(\bm{U}^\top\bm{F}_t)
		\,\text{,}
	\end{split}
\end{equation}

\noindent where $Tr(\cdot)$ denotes the trace of matrix. $\bm{C}_t$ and $\bm{F}_t$ are updated as

\begin{equation}\label{eq.21}
	\begin{split}
		&\bm{C}_t\leftarrow\bm{C}_{t-1}+\bm{v}_t^{'}\bm{v}_t^{'\top} \\
		&\bm{F}_t\leftarrow\bm{F}_{t-1}+(\bm{d}_t^{'}-\bm{s}_t^{'})\bm{v}_t^{'\top}\,\text{,}
	\end{split}
\end{equation}

\noindent where $\bm{C}_0=\bm{0}$, $\bm{F}_0=\bm{0}$, $\bm{d}_t^{'}=diag(\bm{w}_t)\bm{d}_t$, $\bm{s}_t^{'}=diag(\bm{w}_t)\bm{s}_t$, and $\bm{v}_t^{'}=(\bm{U}^\top \bm{U})^{-1}\bm{U}^\top diag(\bm{w}_t)\bm{Uv}_t$. In practice, the $i$-th column $\bm{u}_i$ of $\bm{U}$ can be updated individually while keeping other columns fixed \cite{bertsekas1999nonlinear} as

\begin{equation}\label{eq.22}
	\bm{u}_i^{(t)}\leftarrow\bm{u}_i^{(t-1)}+\frac{1}{\tilde{\bm{c}}_{i,i}}
	(\bm{f}_i-\bm{U}^{(t-1)}\tilde{\bm{c}}_{i})\,\text{,}
\end{equation}

\noindent where $\tilde{\bm{C}}=\bm{C}_t+\eta\sigma_w^2\bm{I}$ and $\tilde{\bm{c}}_i$ and $\bm{f}_i$ are the $i$-th column of $\tilde{\bm{C}}$ and $\bm{F}_t$, respectively. Before foreground extraction, we initialize the background by taking the median of pixel values of the first several frames. Then, the basis $\bm{U}_0$ can be initialized by the bilateral random projection method proposed in \cite{zhou2012bilateral} as

\begin{equation}\label{eq.23}
	\bm{U}^{(0)}=\bm{A}_1(\bm{R}_2^{\top}\bm{A}_1)^{-1}\bm{A}_2^\top
	\,\text{,}
\end{equation}

\noindent where $\bm{R}_1\in\mathbb{R}^{n \times r}$ and $\bm{R}_2\in\mathbb{R}^{m \times r}$ denote two bilateral Gaussian random projections, $\bm{A}_1=\bm{AR}_1$, $\bm{A}_2=\bm{A}^{\top}\bm{R}_2$ and $\bm{A}\in\mathbb{R}^{m \times n}$ denotes matrix of the initial background. In summary, the proposed Hyper RPCA is summarized in Algorithm \ref{alg:1}.

\begin{table*}[htbp]
	\centering
	\caption{Quantitative background extraction results in terms of \textit{PSNR} (in dB) and \textit{SSIM} on 10 sequences.}
	\label{tab.1}
	\renewcommand\tabcolsep{1.6pt}
	\renewcommand\arraystretch{1.2}
	\scriptsize
	\begin{tabularx}{0.97\textwidth}{ccccccccccc}
		\toprule[1pt]
		{}&Corridor&DRoom&Library&Canoe&WSurface&
		Skating&Blizzard&TunnelExit&TimeOfDay&Fountain02 \\
		\midrule													
		{GoDec}&(26.53/0.96)&(26.35/0.95)&(20.20/0.90)&(24.20/0.63)&(26.97/0.80)&
		(20.78/0.96)&(46.65/0.99)&(41.97/0.99)&(39.91/0.99)&(33.52/0.93) \\
		{GoDec+}&(26.97/0.96)&(26.39/0.95)&(19.05/0.89)&(24.24/0.63)&(25.59/0.79)&
		(21.08/0.96)&(47.19/0.99)&(43.10/0.99)&(39.89/0.99)&(32.95/0.93) \\
		{GRASTA}&(30.82/0.90)&(23.19/0.86)&(19.44/0.75)&(21.25/0.55)&(22.50/0.71)&
		(20.52/0.84)&(37.90/0.95)&(34.59/0.95)&(33.74/0.94)&(27.14/0.86) \\
		{incPCP}&(33.57/0.95)&(28.76/0.94)&(23.68/0.90)&(23.28/0.57)&(27.12/0.77)&
		(24.74/0.95)&(45.39/0.99)&(41.54/0.98)&(40.40/0.98)&(32.33/0.91) \\
		{noncvx-RPCA}&(32.31/0.95)&(26.40/0.95)&(18.49/0.87)&(24.24/0.63)&
		(27.93/$\bm{0.82}$)&(23.72/0.93)&($\bm{47.85}$/$\bm{0.99}$)&
		(42.44/$\bm{0.99}$)&(40.30/0.99)&($\bm{33.93}$/$\bm{0.93}$) \\
		{ORPCA}&(36.07/0.94)&(28.65/0.93)&(26.93/0.89)&(26.08/0.65)&(29.09/0.76)&
		(26.63/0.80)&(44.39/0.97)&(36.13/0.96)&(38.37/0.97)&(29.90/0.89) \\
		{PRMF}&(36.70/0.91)&(29.74/0.96)&(16.01/0.69)&(21.09/0.62)&(24.94/0.74)&
		(24.38/0.86)&(35.23/0.84)&(38.51/0.97)&($\bm{45.16}$/0.95)&(33.23/0.93) \\
		{Hyper RPCA}&($\bm{46.59}$/$\bm{0.99}$)&($\bm{44.98}$/$\bm{0.98}$)&
		($\bm{44.21}$/$\bm{0.98}$)&($\bm{26.38}$/$\bm{0.70}$)&($\bm{29.84}$/0.78)&
		($\bm{38.45}$/$\bm{0.97}$)&(46.43/0.99)&($\bm{43.92}$/0.99)&(40.68/$\bm{0.99}$)&
		(33.28/0.92) \\
		\bottomrule[1pt]
	\end{tabularx}
\end{table*}

\begin{figure*}[htpb]
	\centering
	\includegraphics[width=.80\linewidth]{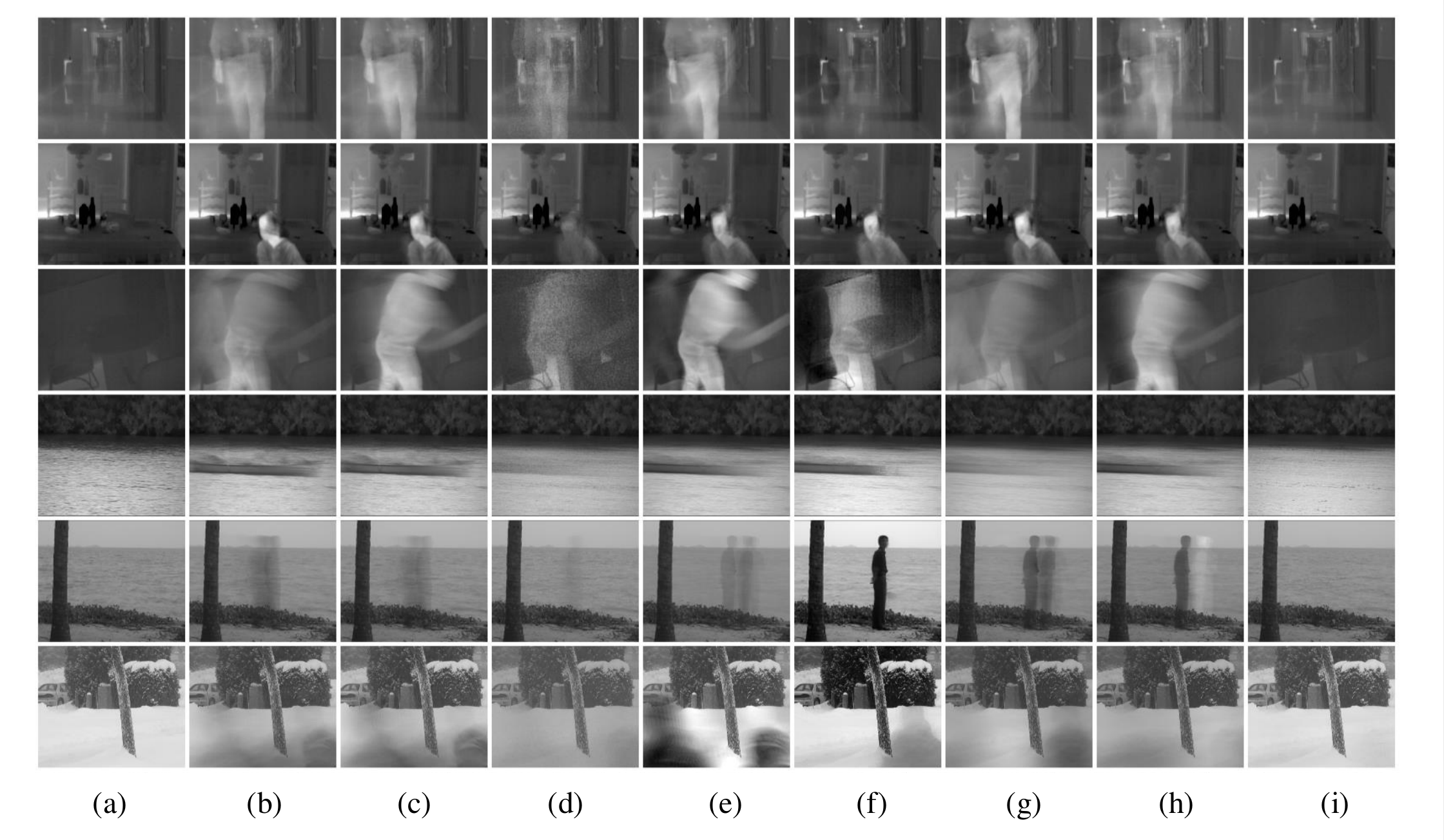}
	\caption{Visual results of different methods for background extraction. (a) Background ground-truth. The background extraction results generated by (b) GRASTA \cite{he2012incremental}, (c) incPCP \cite{rodriguez2016incremental}, (d) ORPCA \cite{feng2013online}, (e) noncvx-RPCA \cite{kang2015robust}, (f) PRMF \cite{wang2012probabilistic}, (g) GoDec \cite{zhou2011godec}, (h) GoDec+ \cite{guo2017godec+} and (i) the proposed Hyper RPCA. From top to bottom: the 0640th frame of Corridor, the 0899th frame of DRoom, the 1020th frame of Library, the 0959th frame of Canoe, the 1525th frame of WSurface and the 1670th frame of Skating.}
	\label{fig.3}
\end{figure*}

% ----------------------------experiments------------------------------

\section{Experiments}\label{sec.experiments}
The major parameters of the proposed Hyper RPCA were set as, $r=25$ and $\sigma=1 \times 10^{+3}$. $\sigma_w^2$ was set in the range of $[1,10] \times 10^{-5}$ according to the differences of pixel intensity between the foreground and background.

We tested the proposed Hyper RPCA on 16 representative video sequences, including 11 challenging clips selected from CDnet \cite{wang2014cdnet} (Canoe, Fountain02 and Overpass sequences in ``Dynamic Background'' category, Boulevard and Traffic sequences in ``Camera Jitter'' category, Corridor, DiningRoom (DRoom) and Library sequences in ``Thermal'' category, Blizzard and Skating sequences in ``Bad Weather'' category, TunnelExit\_0\_35fps (TunnelExit) sequence in ``Low Framerate'' category), 3 sequences (Hall, Lobby and WaterSurface (WSurface)) from I2R \cite{li2004statistical} and 2 sequences (ForegroundAperture (FAperture) and TimeOfDay) from Wallflower \cite{toyama1999wallflower} dataset, respectively. Since the offline methods require all the consecutive frames, which results in a large computational cost, only 200 consecutive frames from each test video were chosen for our experiment.

The proposed method was implemented on Matlab platform. For all of the competing methods, we used the publicly available codes from their official websites, with the default parameters. All the experiments in this paper were performed on a PC with 2.3GHZ Intel Core i5 processor and 8GB of RAM.

\subsection{Experimental Results on Background extraction}
To verify the performance of background extraction, the proposed Hyper RPCA was compared with 7 state-of-art-methods. The utilized offline methods include: noncvx-RPCA \cite{kang2015robust}, PRMF \cite{wang2012probabilistic}, GoDec \cite{zhou2011godec} and GoDec+ \cite{guo2017godec+}, and online methods include: incPCP \cite{rodriguez2016incremental}, GRASTA \cite{he2012incremental}, ORPCA \cite{feng2013online}. Peak signal-to-noise ratio (\textit{PSNR}) and structural similarity index (\textit{SSIM}) \cite{wang2004image} were used as the quantitative metrics for the background extraction result. Ground-truth background images of static videos are obtained by averaging all background frames which exclude the foreground part. Table \ref{tab.1} shows the average \textit{PSNR} and \textit{SSIM} values of different methods for some video sequences without camera jitter. (In this paper, bold number shows the best result in each Table.) The proposed Hyper RPCA achieves highest scores in 7 sequences in terms of \textit{PSNR} and 6 sequences in terms of \textit{SSIM}, and competitive performance in the rest.

\begin{table*}[htbp]
	\centering
	\caption{Quantitative foreground detection results in terms of \textit{F-measure} on 12 sequences.}
	\label{tab.2}
	\renewcommand\tabcolsep{4.9pt}
	\renewcommand\arraystretch{1.2}
	\scriptsize
	\begin{tabularx}{0.97\textwidth}{cccccccccccccc}
		\toprule[1pt]
		{}&Blizzard&Corridor&DRoom&TunnelExit&Library&TimeOfDay&Boulevard&
		Canoe&Fountain02&Skating&Traffic&WSurface&Average \\
		\midrule
		{SuBSENSE}&$\bm{0.97}$&0.98&0.93&0.83&0.98&
		0.78&0.76&0.80&0.89&0.96&0.44&0.91&0.85 \\
		{PAWCS}&0.77&0.97&0.95&$\bm{0.86}$&0.98&
		0.44&0.57&0.86&0.91&0.97&0.46&0.85&0.80 \\
		{WeSamBE}&0.76&0.75&0.61&0.77&0.66&
		$\bm{0.85}$&0.73&0.72&0.81&0.75&0.75&0.85&0.75 \\
		{COROLA}&0.90&0.98&0.94&0.84&0.99&
		0.66&0.81&$\bm{0.87}$&$\bm{0.92}$&0.97&0.74&$\bm{0.92}$&0.88 \\
		{OPRMF}&0.86&0.85&0.86&0.71&0.84&
		0.42&0.75&0.73&0.85&0.82&0.73&0.83&0.77 \\
		{SOBS}&0.61&0.96&0.90&0.38&0.99&
		0.51&0.45&0.71&0.81&0.96&0.58&0.87&0.73 \\
		{ViBe}&0.58&0.94&0.83&0.53&0.98&
		0.41&0.35&0.52&0.68&0.95&0.57&0.86&0.68 \\
		{GMM}&0.90&0.72&0.49&0.63&0.50&
		0.56&0.41&0.42&0.68&0.82&0.58&0.41&0.59 \\
		{DECOLOR}&0.78&0.96&0.94&0.59&0.98&
		0.51&$\bm{0.93}$&0.41&0.81&$\bm{0.98}$&0.79&0.84&0.79 \\
		{RegL1}&0.94&0.93&0.73&0.76&0.94&
		0.58&0.74&0.72&0.85&0.95&0.75&0.90&0.81 \\
		{MAMR}&0.94&0.93&0.73&0.76&0.92&
		0.58&0.74&0.70&0.85&0.95&0.75&0.90&0.80 \\
		{MoG-RPCA}&0.94&0.83&0.70&0.76&0.66&
		0.58&0.76&0.59&0.85&0.88&0.76&0.82&0.76 \\
		{Hyper RPCA}&0.94&$\bm{0.99}$&$\bm{0.96}$&0.76&$\bm{0.99}$&
		0.77&0.88&0.76&0.89&0.97&$\bm{0.84}$&0.87&$\bm{0.89}$ \\
		\bottomrule[1pt]
	\end{tabularx}
\end{table*}

\begin{figure*}[htpb]
	\centering
	\includegraphics[width=.95\linewidth]{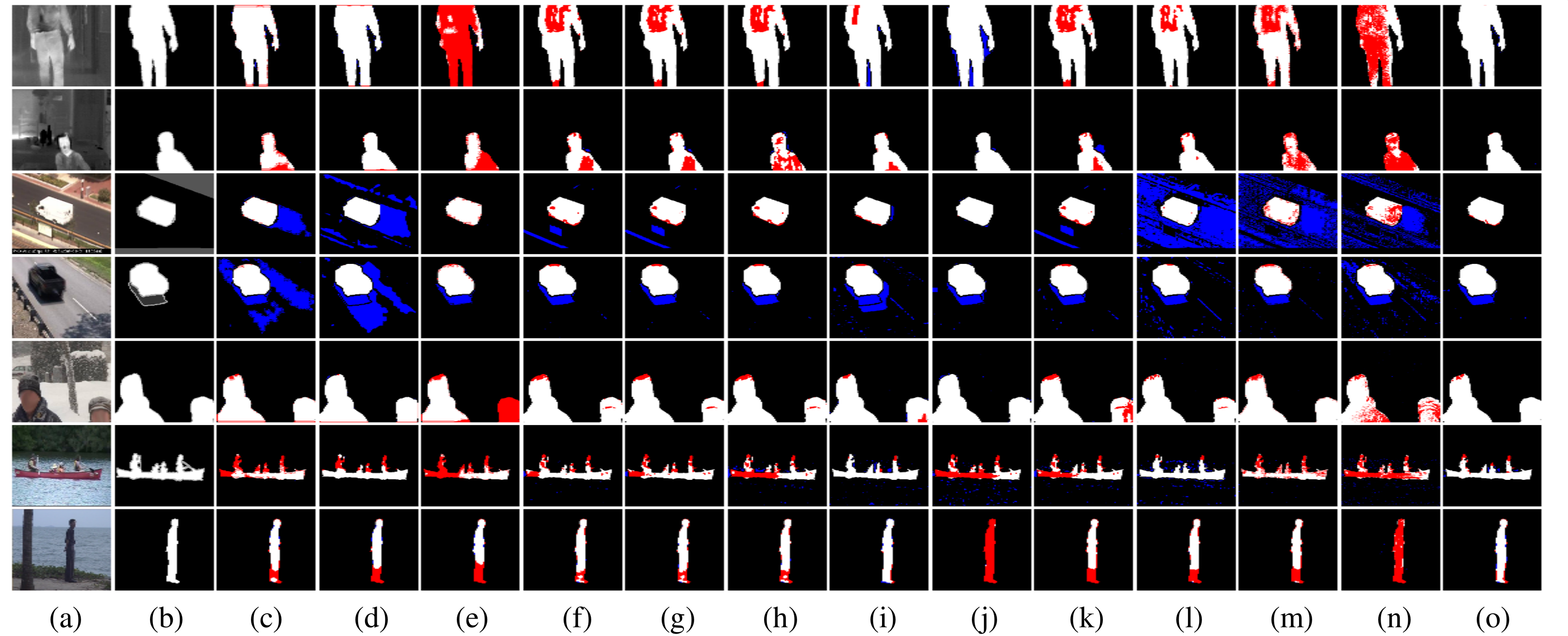}
	\caption{Visual results of different methods for foreground detection. (a) Input images. (b) Ground-truth foreground mask. The foreground detection results generated by (c) SuBSENSE \cite{st2014subsense}, (d) PAWCS \cite{st2016universal}, (e) WeSamBE \cite{jiang2017wesambe}, (f) RegL1 \cite{zheng2012practical}, (g) MAMR \cite{ye2015foreground}, (h) MoG-RPCA \cite{zhao2014robust}, (i) COROLA \cite{shakeri2016corola}, (j) DECOLOR \cite{zhou2012moving}, (k) OPRMF \cite{wang2012probabilistic}, (l) SC-SOBS \cite{maddalena2012sobs}, (m) ViBe \cite{barnich2010vibe}, (n) GMM \cite{zivkovic2004improved} and (o) the proposed Hyper RPCA. From top to bottom: the 0631th frame of Corridor, the 0877th frame of DRoom, the 1210th frame of Boulevard, the 0968th frame of Traffic, the 1667th frame of Skating, the 0957th frame of Canoe and the 1594th frame of WSurface. (Write represents correctly detected foreground, red represents missing pixels, and blue represents false alarm pixels.)}
	\label{fig.4}
\end{figure*}

Several representative background extraction visual results from the most challenging test videos are demonstrated in Fig. \ref{fig.3}, from which we can see that the results of the proposed Hyper RPCA have superior performance over other methods. The extracted backgrounds by the proposed Hyper RPCA are obviously closer to the ground-truth, while other methods produce ghosting artifacts in different degrees.

Since the ground-truth is difficult to estimate for Canoe and Fountain02, which have dynamic background, the \textit{PSNR} and \textit{SSIM} values of different methods are close. For Blizzard, TunnelExit, TimeOfDay and Fountain02, the moving objects run fast in the scenes, all the methods can estimate the background well. However, when it comes to moving slowly objects, for example, the walking men in Corridor and running boats in Canoe, which move through an area of frame and shade background for a long time, the results from other methods contain severe ghosting artifacts. Our method generates much cleaner results than others, which demonstrates the potential of the proposed method for this situation.

\subsection{Experimental Results on Moving object detection}
While the background is extracted, the moving objects can be obtained via background subtraction. To demonstrate the effectiveness of the proposed Hyper RPCA, we compared our method with 12 state-of-the-art methods for foreground detection, including four offline methods: RegL1 \cite{zheng2012practical}, MAMR \cite{ye2015foreground}, MoG-RPCA \cite{zhao2014robust}, DECOLOR \cite{zhou2012moving}, and eight online methods: SuBSENSE \cite{st2014subsense}, PAWCS \cite{st2016universal}, WeSamBE \cite{jiang2017wesambe}, COROLA \cite{shakeri2016corola}, OPRMF \cite{wang2012probabilistic}, SC-SOBS \cite{maddalena2012sobs}, ViBe \cite{barnich2010vibe}, GMM \cite{zivkovic2004improved}. The foreground detection result is assessed using \textit{F-measure} which is defined as

\begin{equation}\label{eq.24}
	\textit{F-measure}=2 \times 
	\frac{\textit{precision} \times \textit{recall}}
	{\textit{precision} + \textit{recall}}
	\,\text{.}
\end{equation}

\noindent where \textit{precision}=\textit{TP}/(\textit{TP}+\textit{FP}) and \textit{recall}=\textit{TP}/(\textit{TP}+\textit{FN}). True Positives (\textit{TP}) denotes the number of pixels correctly classified as foreground objects, False Positives (\textit{FP}) represents the number of pixels incorrectly classified as foreground object, and False Negatives (\textit{FN}) is the number of pixels incorrectly classified as background. Table \ref{tab.2} shows the quantitative results of different methods on some test sequences, from which we can see that our proposed Hyper RPCA obtains the highest average \textit{F-measure}. Although in some scenarios, our method did not achieve the best score, the gaps between ours and the first places are inconspicuous.

\begin{table*}[htbp]
	\centering
	\caption{Comparison the average \textit{F-measure} based on the CDnet dataset.}
	\label{tab.3}
	\renewcommand\tabcolsep{3.9pt}
	\renewcommand\arraystretch{1.2}
	\scriptsize
	\begin{tabularx}{0.97\textwidth}{cccccccccccc}
		\toprule[1pt]
		Category&SuBSENSE&SGSM-BS-block&PCP&SC-SOBS&GMM&COROLA&
		DECOLOR&GRASTA&incPCP&OMoGMF+TV&Hyper RPCA \\
		\midrule
		Baseline&$\bm{0.95}$&0.93&0.60&0.92&0.91&0.85&
		0.92&0.66&0.81&0.85&0.94 \\
		Dynamic Background&0.81&0.83&0.34&0.65&0.54&$\bm{0.86}$&
		0.70&0.35&0.71&0.76&0.76 \\
		Camera Jitter&0.81&0.81&0.54&0.82&0.56&0.82&
		0.77&0.43&0.78&0.78&$\bm{0.83}$ \\
		Shadow&$\bm{0.89}$&0.86&0.51&0.82&0.73&0.78&
		0.83&0.52&0.74&0.68&0.87 \\
		Thermal&0.81&$\bm{0.82}$&0.34&0.68&0.65&0.80&
		0.70&0.42&0.70&0.70&0.80 \\
		Intermittent Object Motion&0.65&0.70&0.32&0.59&0.53&0.71&
		0.59&0.35&0.75&0.71&$\bm{0.77}$ \\
		Bad Weather&$\bm{0.86}$&0.80&0.76&0.66&0.54&0.78&
		0.76&0.68&N/A&0.78&0.83 \\
		Low Framerate&0.64&0.73&0.44&0.55&0.50&N/A&
		0.50&N/A&N/A&N/A&$\bm{0.75}$ \\
		\midrule[1pt]
		Average&0.80&0.81&0.48&0.71&0.62&0.80&
		0.72&0.48&0.74&0.75&$\bm{0.81}$ \\
		\bottomrule[1pt]
	\end{tabularx}
\end{table*}

\begin{figure*}[htpb]
	\centering
	\includegraphics[width=0.95\linewidth]{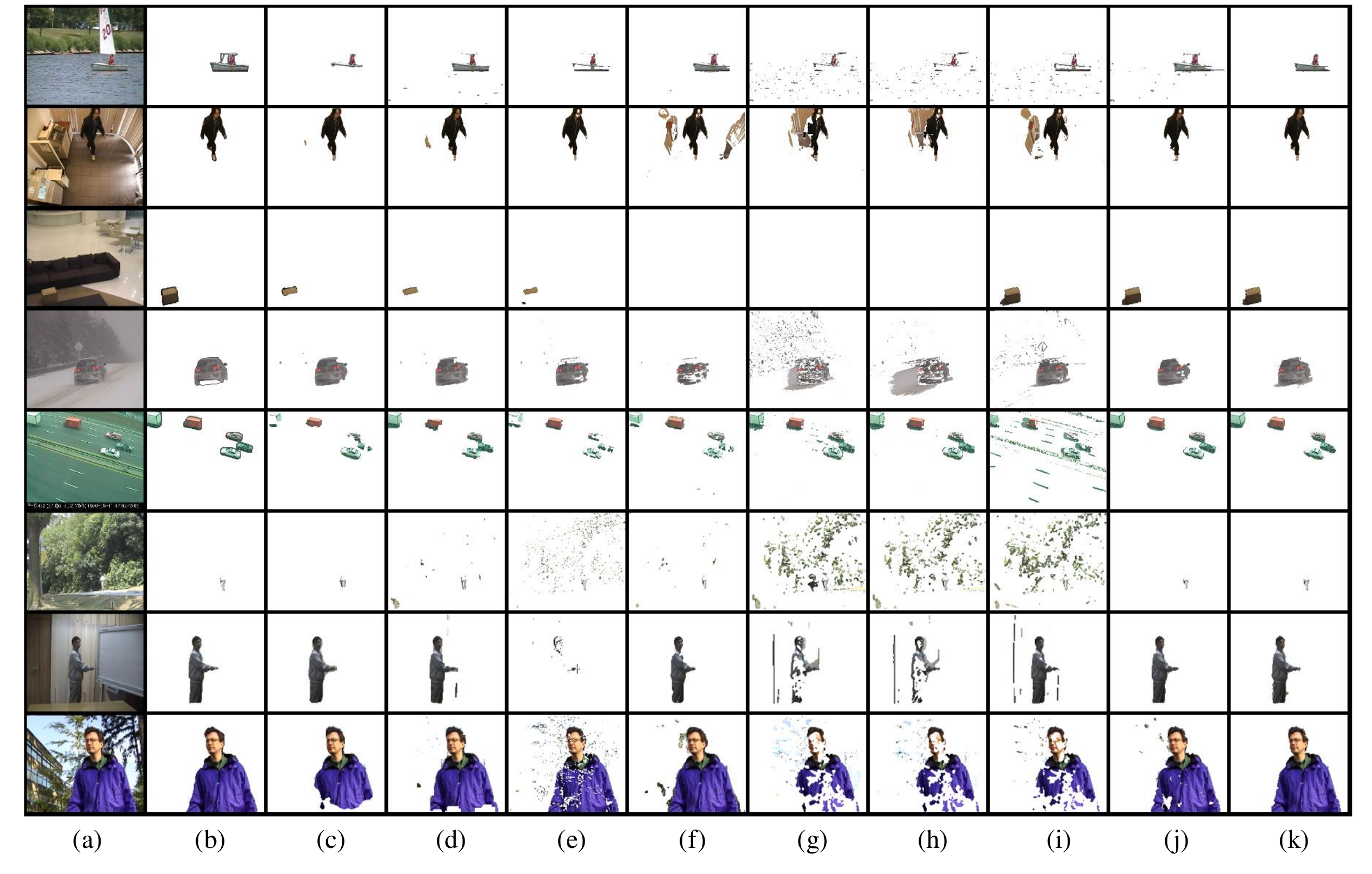}
	\caption{Visual results of different methods on 12 video sequences. (a) Input image frame. (b) Ground-truth image of moving object. Moving objects detection results generated by (c) SuBSENSE \cite{st2014subsense}, (d) COROLA \cite{shakeri2016corola}, (e) GMM \cite{zivkovic2004improved}, (f) SC-SOBS \cite{maddalena2012sobs}, (g) GRASTA \cite{he2012incremental}, (h) incPCP \cite{rodriguez2016incremental}, (i) OMoGMF+TV \cite{yong2017robust}, (j) SGSM-BS-block \cite{shi2018robust} and (k) the proposed Hyper RPCA. From top to bottom: the 7106th frame of Boats, the 2305th frame of CopyMachine, the 500th frame of Sofa, the 846th frame of SnowFall, the 1080th frame of Turnpike, the 1392th frame of Campus, the 23854th frame of Curtain and the 247th frame of WavingTrees.}
	\label{fig.5}
\end{figure*}

Fig. \ref{fig.4} shows several visual results of foreground detection generated by the test methods on some test videos. In Fig. \ref{fig.4}, it can be seen that for indoor scenarios, such as Corridor and DRoom (rows 1 and 2), where people keep moving slowly across many frames, PAWCS, DECOLOR and Hyper RPCA perform better than other methods, and other methods fail to accurately detect the objects due to the pollution of the background. For the scenarios with camera jitter, such as Boulevard and Traffic (rows 3 and 4), whose background changes all the time, only WeSamBE, MOG-RPCA and Hyper RPCA can detect the foreground objects accurately, and in contrast, other methods, such as SuBSENSE and PAWCS, misclassified the background as foreground. For the scenarios with dynamic background, such as Canoe and WSurface (rows 6 and 7), DECOLOR and WeSamBE fail to detect the complete foreground part, in contrast, the shapes of people and boat are complete in the mask extracted by the proposed Hyper RPCA. For the outdoor scenario Skating (row 5), where people keep moving from the right side of the scene to the left, WeSamBE fails to detect the moving people on the right. It can be also noticed that COROLA performed well in each scenario, but the extracted masks are not as accurate as ours and it produced some fake positions in some results. Overall, the proposed Hyper RPCA achieves best visual performance on the test video sequences in all the methods.

\begin{table*}[htbp]
	\centering
	\caption{Comparison of average \textit{F-measure} on I2R and Wallflower dataset.}
	\label{tab.4}
	\renewcommand\tabcolsep{4.8pt}
	\renewcommand\arraystretch{1.2}
	\scriptsize
	\begin{tabularx}{0.97\textwidth}{ccccccccccccc}
		\toprule[1pt]
		Dataset&COROLA&DECOLOR&TVRPCA&SRPCA&GRASTA&incPCP&OMoGMF+TV&3TD&RegL1&MAMR&SGSM-BS-block&Hyper RPCA \\
		\midrule
		I2R&$\bm{0.81}$&0.74&0.69&0.80&0.63&0.62&0.77&0.72&0.63&0.75&0.78&0.79 \\
		Wallflower&0.75&0.59&0.61&0.85&0.33&N/A&0.82&0.75&N/A&0.80&N/A&$\bm{0.86}$ \\
		\bottomrule[1pt]
	\end{tabularx}
\end{table*}

\begin{figure*}[htbp]
	\centering
	\includegraphics[width=.95\linewidth]{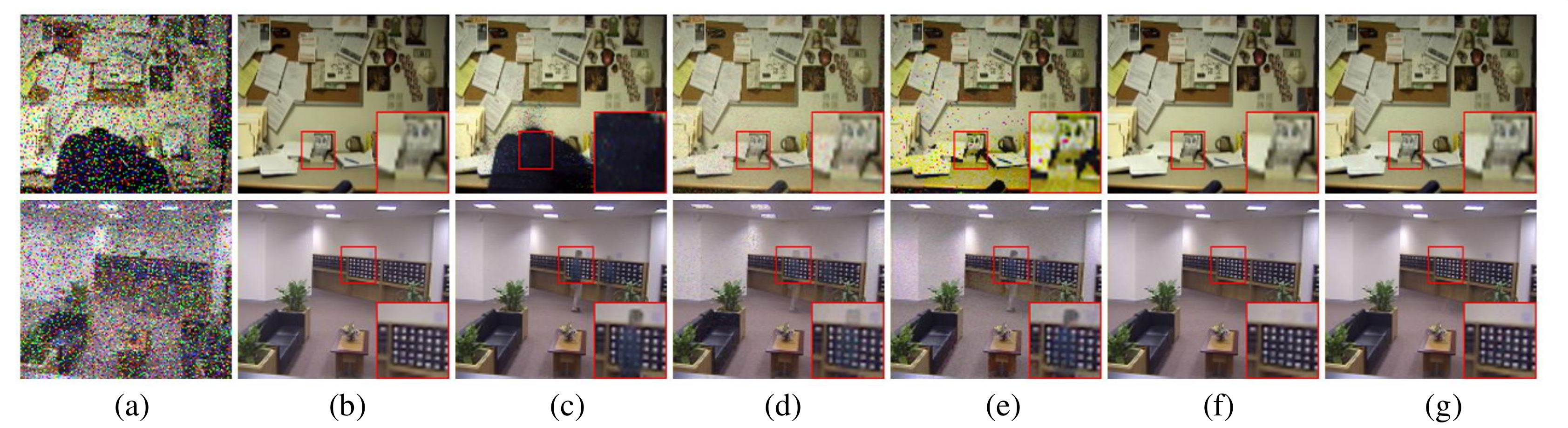}
	\caption{Background extraction results of different methods on test sequences corrupted with non-Gaussian noise. (a) Noisy image. (b) Background ground-truth. The background extraction results generated by (c) PCP \cite{candes2011robust}, (d) ORPCA \cite{feng2013online}, (e) proposed LSM-ORPCA, (f) proposed MCC-ORPCA and (g) proposed Hyper RPCA. From top to bottom: the 945th frame of FAperture and the 2545th frame of Lobby.}
	\label{fig.6}
\end{figure*}

\subsection{Experimental Results on Long Testing sequences}
To further demonstrate the performance of the proposed Hyper RPCA for foreground detection, we compared Hyper RPCA with other methods on three long sequence datasets, including CDnet \cite{wang2014cdnet}, I2R \cite{li2004statistical} and Wallflower \cite{toyama1999wallflower}. The comparison methods include six offline methods: 3TD \cite{oreifej2012simultaneous}, PCP \cite{candes2011robust}, DECOLOR \cite{zhou2012moving}, RegL1 \cite{zheng2012practical}, TVRPCA \cite{cao2015total}, MAMR \cite{ye2015foreground}, and night online methods: SuBSENSE \cite{st2014subsense}, GMM \cite{zivkovic2004improved}, SC-SOBS \cite{maddalena2012sobs}, COROLA \cite{shakeri2016corola}, GRASTA \cite{he2012incremental}, incPCP \cite{rodriguez2016incremental}, OMoGMF+TV \cite{yong2017robust}, SRPCA \cite{javed2016spatiotemporal}, SGSM-BS-block \cite{shi2018robust}.

Eight categories, including ``Baseline'', ``Dynamic Background'', ``Camera Jitter'', ``Shadow'', ``Thermal'', ``Intermittent Object Motion'', ``Bad Weather'' and ``Low Framerate'', from CDnet 2014 dataset were tested. Table \ref{tab.3} shows the quantitative results of all the methods. (In this paper, N/A indicates that the authors did not report the performance for these categories or datasets in their original references.) As shown in Table \ref{tab.3}, Hyper RPCA outperforms most of other methods and is competitive with SGSM-BS-block \cite{shi2018robust}, SuBSENSE \cite{st2014subsense}, WeSamBE \cite{jiang2017wesambe} and COROLA \cite{shakeri2016corola} methods, which can effectively deal with complex scenes in practice. I2R and Wallflower datasets consist of nine and six videos with complex background respectively. As shown in Table \ref{tab.4}, Hyper RPCA achieves the best performance on the Wallflower dataset on average. The performance of Hyper RPCA on the I2R dataset outperforms the other methods, except COROLA and SRPCA, which are state-of-the-art methods for moving object detection.

The visual results of eight typical methods and Hyper RPCA are shown in Fig. \ref{fig.5}. In Fig. \ref{fig.5}, it can be seen that for scenario Sofa (row 3), where box was kept in a fixed place for a long time, only OMoGMF+TV, SGSM-BS-block and Hyper RPCA can detect foreground accurately, and other methods fail to detect the box wholly. For indoor scenario CopyMachine (row 2) and outdoor scenario Campus (row 6), GRASTA, incPCP and OMoGMF+TV misclassified the background as foreground. It can be noticed that the shape of people, for scenario WavingTrees (row 8), are complete in the mask extracted by Hyper RPCA, and other methods produced some fake positions in results. Overall, the foreground detection results of Hyper RPCA are closer to the ground-truth images.

\begin{figure*}[htbp]
	\centering
	\subfloat[]{
		\begin{minipage}[]{0.3\linewidth}
			\centering
			\includegraphics[width=5cm]{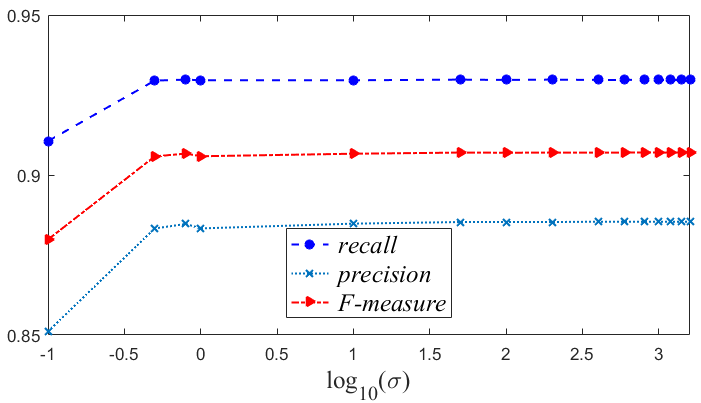}
			%\caption{fig1}
		\end{minipage}%
	}%
	\subfloat[]{
		\begin{minipage}[]{0.3\linewidth}
			\centering
			\includegraphics[width=5cm]{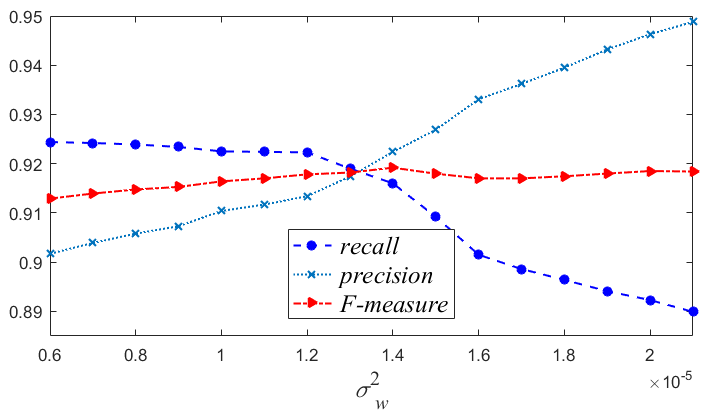}
			%\caption{fig1}
		\end{minipage}%
	}%
	\subfloat[]{
		\begin{minipage}[]{0.3\linewidth}
			\centering
			\includegraphics[width=5cm]{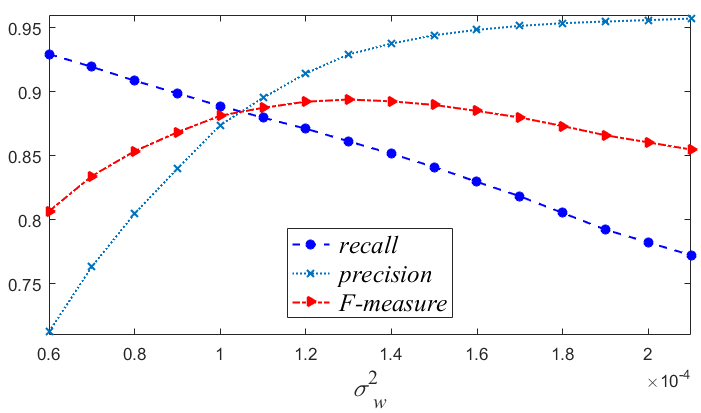}
			%\caption{fig1}
		\end{minipage}%
	}%
	\caption{The average \textit{F-measure}, \textit{precision} and \textit{recall} curves as functions of (a) the Gaussian kernel $\sigma$ in log domain, (b) the variance of the Gaussian error $\sigma_w^2$ of the sequences with similar spatial homogeneity between foreground and background and (c) $\sigma_w^2$ of the sequences with dissimilar spatial homogeneity between foreground and background.}
	\label{fig.7}
\end{figure*}

\begin{figure*}[htbp]
	\centering
	\subfloat[]{
		\begin{minipage}[]{0.5\linewidth}
			\centering
			\includegraphics[width=8cm]{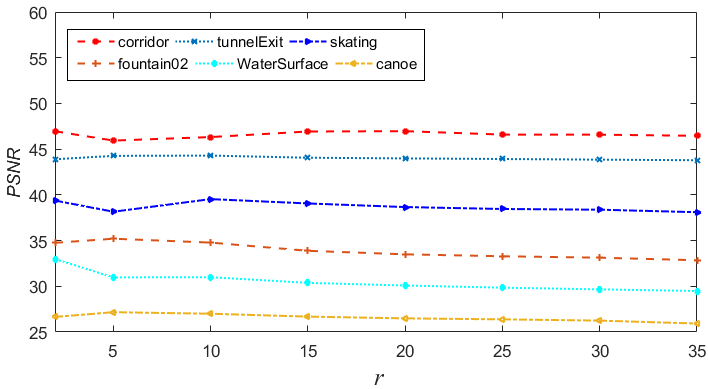}
			%\caption{fig1}
		\end{minipage}%
	}%
	\subfloat[]{
		\begin{minipage}[]{0.5\linewidth}
			\centering
			\includegraphics[width=8cm]{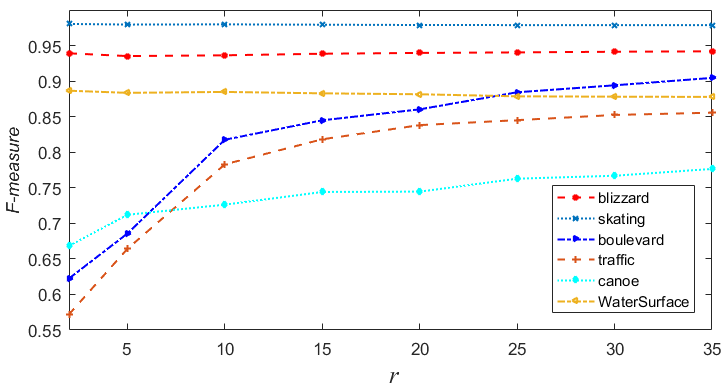}
			%\caption{fig1}
		\end{minipage}%
	}%
	\caption{The average \textit{PSNR} and \textit{F-measure} curves as functions of the bilateral random projections $r$. (a) The \textit{PSNR} curve as function of $r$ and (b) The \textit{F-measure} curve as function of $r$.}
	\label{fig.8}
\end{figure*}

\subsection{Ablation Study}
To verify the effectiveness of the proposed MCC and LSM regularization terms, we implemented three variants of the proposed model, i.e., the LSM-based method without MCC (denoted as LSM-ORPCA), the MCC-based method without LSM (denoted as MCC-ORPCA), and Hyper RPCA based on LSM and MCC. Some representative results are shown in Table \ref{tab.5}, from which we can see that LSM-ORPCA and Hyper RPCA significantly outperforms PCP and ORPCA. By utilizing MCC to model the error part, Hyper RPCA performs better than LSM-ORPCA. MCC-ORPCA, which uses $\ell_1$-norm to model foreground pixels, degrades while dealing with dynamic backgrounds. Experiment results demonstrate that LSM model has the advantage of characterizing the varying sparsity of the foreground pixels, and is more suitable for foreground modeling than $\ell_1$-norm in practice.

\begin{table}[t]
	\centering
	\caption{Quantitative foreground detection results in terms of \textit{F-measure} on 6 sequences.}
	\label{tab.5}
	\renewcommand\tabcolsep{2pt}
	\renewcommand\arraystretch{1.2}
	\scriptsize
	\begin{tabularx}{0.47\textwidth}{cccccccc}
		\toprule[1pt]
		{}&Corridor&DRoom&Library&Boulevard&Skating&WSurface&Average \\
		\midrule
		PCP&0.74&0.79&0.69&0.85&0.75&0.60&0.73 \\
		ORPCA&0.89&0.78&0.84&0.80&0.73&0.74&0.79 \\
		MCC-ORPCA&0.99&0.96&0.97&0.52&0.97&$\bm{0.90}$&0.88 \\
		LSM-ORPCA&0.99&$\bm{0.96}$&0.99&0.86&0.97&0.87&0.94 \\
		Hyper RPCA&$\bm{0.99}$&0.96&$\bm{0.99}$&$\bm{0.88}$&
		$\bm{0.97}$&0.87&$\bm{0.94}$ \\
		\bottomrule[1pt]
	\end{tabularx}
\end{table}

\begin{table}[t]
	\centering
	\caption{Quantitative background extraction results in terms of \textit{PSNR} (in dB) and \textit{SSIM} on 4 noisy sequences.}
	\label{tab.6}
	\renewcommand\tabcolsep{3.2pt}
	\renewcommand\arraystretch{1.2}
	\scriptsize
	\begin{tabularx}{0.47\textwidth}{ccccc}
		\toprule[1pt]
		{}&FAperture&Hall&Lobby&Overpass \\
		\midrule
		PCP&(14.18/0.77)&(27.43/0.93)&(28.98/0.95)&(24.14/0.86) \\
		ORPCA&(20.29/0.90)&(23.77/0.87)&(28.37/0.90)&(24.43/0.83) \\
		LSM-ORPCA&(19.73/0.83)&(25.22/0.89)&(27.84/0.85)&(25.78/0.84) \\
		MCC-ORPCA&(25.12/$\bm{0.97}$)&($\bm{32.25}$/$\bm{0.96}$)&
		(31.88/0.97)&($\bm{28.83}$/$\bm{0.91}$) \\
		Hyper RPCA&($\bm{29.14}$/0.97)&(31.83/0.96)&
		($\bm{32.70}$/$\bm{0.97}$)&(28.61/0.91) \\
		\bottomrule[1pt]
	\end{tabularx}
\end{table}

In Table \ref{tab.6} and Fig. \ref{fig.6}, the results demonstrate that MCC can handle non-Gaussian noise well. MCC-ORPCA can gain cleaner background than PCP and ORPCA. Original test sequences were corrupted by Poisson+Salt\&Pepper noise (corruption percentage is set to 20\%). By introducing LSM model, Hyper RPCA obtains the higher average \textit{PSNR} (30.57dB) and \textit{SSIM} (0.9580) values than MCC-ORPCA (29.52dB/0.9579). Similar to PCP and ORPCA, the background extraction results of LSM-ORPCA have few noise and artifacts.

\subsection{Parameters Selection}
In the proposed method, three major parameters need to be tuned, including the setting of Gaussian kernel function $\sigma$, the bilateral random projections number $r$ and the variance of the Gaussian error $\sigma_w^2$. Fig. \ref{fig.7} shows the average \textit{F-measure}, precision and recall curves as functions of $\log_{10}(\sigma)$ and $\sigma_w^2$ on 12 test video sequences used in Table \ref{tab.2}. From Fig. \ref{fig.7} (a), we can see that the performance of the proposed method is insensitive to $\sigma$. From Fig. \ref{fig.7} (b) and (c), we can see that the precision result increases and the recall result decreases with the value of $\sigma_w^2$ increases. In our implementation, we set $\sigma=1 \times 10^{+3}$ experimentally. $\sigma_w^2=1 \times 10^{-5}$ for the sequences with similar spatial homogeneity between foreground and background, or $\sigma_w^2=1 \times 10^{-4}$ for the opposite situation.

Fig. \ref{fig.8} shows the \textit{PSNR} and \textit{F-measure} curves as functions of $r$ on 6 test video sequences. From Fig. \ref{fig.8} (a), we can see that the result of background extraction is insensitive to $r$. From Fig. \ref{fig.8} (b), it can be observed that as $r$ increases the performance is improved for the sequences with dynamic background. On the other hand, the performance for the sequences with stable background is insensitive to $r$. Considering the time cost and accuracy of foreground detection, we set $r=25$ in our experiments.

\subsection{Computational Complexity}
The computational complexity of the proposed method mainly depends on the costs of the optimization schemes, including 1) estimating $\bm{w}_t$; 2) estimating $\bm{v}_t$; 3) estimating $\bm{s}_t$; and 4) updating $\bm{U}$. During one iteration per frame, the complexities of estimating $\bm{w}_t$, $\bm{v}_t$, $\bm{s}_t$ and $\bm{U}$ are $O(p)$, $O(pr^2)$, $O(p)$ and $O(pr^2)$ respectively. Thus, the total time complexity of the proposed online algorithm for a video sequence with $T$ frames is $O(Tpr^2)$, which is linearly proportional to the size and number of the frames. We also compared the performing time of the proposed method with some representative methods, including SuBSENSE \cite{st2014subsense}, PAWCS \cite{st2016universal}, WeSamBE \cite{jiang2017wesambe}, DECOLOR \cite{zhou2012moving}, MoG-RPCA \cite{zhao2014robust}, and ORPCA \cite{feng2013online}. Table \ref{tab.7} reports the average running time of different methods for 12 test video sequences, and the proposed method is quite competitive in running time.

\begin{table}[htbp]
	\centering
	\caption{Time costs for Foreground Detection \\ (seconds per frame).}
	\label{tab.7}
	\renewcommand\tabcolsep{1.8pt}
	\renewcommand\arraystretch{1.2}
	\scriptsize
	\begin{tabularx}{0.47\textwidth}{ccccccc}
		\toprule[1pt]
		SuBSENSE&PAWCS&WeSamBE&DECOLOR&MoG-RPCA&ORPCA&Hyper RPCA \\
		\midrule
		3.45&0.64&5.78&2.29&1.25&3.97&1.10 \\
		\bottomrule[1pt]
	\end{tabularx}
\end{table}

% ----------------------------conclusion------------------------------

\section{Conclusion}\label{sec.conclusion}
Although RPCA-based models have been successfully used for moving object detection, the intrinsic shortcomings of the $\ell_1$-norm and $\ell_2$-norm still need to be solved. In this paper, we proposed an online background subtraction model by integrating maximum correntropy criterion (MCC) and Laplacian scale mixture (LSM) models for moving object detection. The proposed Hyper RPCA applies correntropy as the error measurement to improve the robustness to non-Gaussian modeling error. Moreover, the LSM model is adopted to formulate foreground pixels of the moving object. Experimental results show that our algorithm has competitive performance to the state-of-the-art moving object detection methods and demonstrates powerful abilities in characterizing non-Gaussian errors and the sparsity of sparse foreground pixels. In the future, developing an adaptive parameter selection strategy for different complex scenarios will be considered.

% ----------------------------references------------------------------

\footnotesize
\bibliographystyle{IEEEtran}
\bibliography{reference}

\begin{thebibliography}{10}
\providecommand{\url}[1]{#1}
\csname url@samestyle\endcsname
\providecommand{\newblock}{\relax}
\providecommand{\bibinfo}[2]{#2}
\providecommand{\BIBentrySTDinterwordspacing}{\spaceskip=0pt\relax}
\providecommand{\BIBentryALTinterwordstretchfactor}{4}
\providecommand{\BIBentryALTinterwordspacing}{\spaceskip=\fontdimen2\font plus
\BIBentryALTinterwordstretchfactor\fontdimen3\font minus
  \fontdimen4\font\relax}
\providecommand{\BIBforeignlanguage}[2]{{%
\expandafter\ifx\csname l@#1\endcsname\relax
\typeout{** WARNING: IEEEtran.bst: No hyphenation pattern has been}%
\typeout{** loaded for the language `#1'. Using the pattern for}%
\typeout{** the default language instead.}%
\else
\language=\csname l@#1\endcsname
\fi
#2}}
\providecommand{\BIBdecl}{\relax}
\BIBdecl

\bibitem{paul2010video}
M.~Paul, W.~Lin, C.~T. Lau, and B.-s. Lee, ``Video coding using the most common
  frame in scene,'' in \emph{2010 IEEE International Conference on Acoustics,
  Speech and Signal Processing}.\hskip 1em plus 0.5em minus 0.4em\relax IEEE,
  2010, pp. 734--737.

\bibitem{chen2012surveillance}
C.~Chen, J.~Cai, W.~Lin, and G.~Shi, ``Surveillance video coding via low-rank
  and sparse decomposition,'' in \emph{Proceedings of the 20th ACM
  international conference on Multimedia}, 2012, pp. 713--716.

\bibitem{jodoin2012behavior}
P.-M. Jodoin, V.~Saligrama, and J.~Konrad, ``Behavior subtraction,'' \emph{IEEE
  transactions on image processing}, vol.~21, no.~9, pp. 4244--4255, 2012.

\bibitem{cullen2012detection}
D.~Cullen, J.~Konrad, and T.~D. Little, ``Detection and summarization of
  salient events in coastal environments,'' in \emph{2012 IEEE Ninth
  International Conference on Advanced Video and Signal-Based
  Surveillance}.\hskip 1em plus 0.5em minus 0.4em\relax IEEE, 2012, pp. 7--12.

\bibitem{yilmaz2006object}
A.~Yilmaz, O.~Javed, and M.~Shah, ``Object tracking: A survey,'' \emph{Acm
  computing surveys (CSUR)}, vol.~38, no.~4, pp. 13--es, 2006.

\bibitem{oreifej2012simultaneous}
O.~Oreifej, X.~Li, and M.~Shah, ``Simultaneous video stabilization and moving
  object detection in turbulence,'' \emph{IEEE transactions on pattern analysis
  and machine intelligence}, vol.~35, no.~2, pp. 450--462, 2012.

\bibitem{xu2013gosus}
J.~Xu, V.~K. Ithapu, L.~Mukherjee, J.~M. Rehg, and V.~Singh, ``Gosus:
  Grassmannian online subspace updates with structured-sparsity,'' in
  \emph{Proceedings of the IEEE International Conference on Computer Vision},
  2013, pp. 3376--3383.

\bibitem{gao2014block}
Z.~Gao, L.-F. Cheong, and Y.-X. Wang, ``Block-sparse rpca for salient motion
  detection,'' \emph{IEEE transactions on pattern analysis and machine
  intelligence}, vol.~36, no.~10, pp. 1975--1987, 2014.

\bibitem{xin2015background}
B.~Xin, Y.~Tian, Y.~Wang, and W.~Gao, ``Background subtraction via generalized
  fused lasso foreground modeling,'' in \emph{Proceedings of the IEEE
  Conference on Computer Vision and Pattern Recognition}, 2015, pp. 4676--4684.

\bibitem{gemignani2016robust}
G.~Gemignani and A.~Rozza, ``A robust approach for the background subtraction
  based on multi-layered self-organizing maps,'' \emph{IEEE transactions on
  image processing}, vol.~25, no.~11, pp. 5239--5251, 2016.

\bibitem{javed2016spatiotemporal}
S.~Javed, A.~Mahmood, T.~Bouwmans, and S.~K. Jung, ``Spatiotemporal low-rank
  modeling for complex scene background initialization,'' \emph{IEEE
  Transactions on Circuits and Systems for Video Technology}, vol.~28, no.~6,
  pp. 1315--1329, 2016.

\bibitem{bouwmans2017decomposition}
T.~Bouwmans, A.~Sobral, S.~Javed, S.~K. Jung, and E.-H. Zahzah, ``Decomposition
  into low-rank plus additive matrices for background/foreground separation: A
  review for a comparative evaluation with a large-scale dataset,''
  \emph{Computer Science Review}, vol.~23, pp. 1--71, 2017.

\bibitem{javed2017background}
S.~Javed, A.~Mahmood, T.~Bouwmans, and S.~K. Jung, ``Background--foreground
  modeling based on spatiotemporal sparse subspace clustering,'' \emph{IEEE
  Transactions on Image Processing}, vol.~26, no.~12, pp. 5840--5854, 2017.

\bibitem{zheng2006extracting}
J.~Zheng, Y.~Wang, N.~L. Nihan, and M.~E. Hallenbeck, ``Extracting roadway
  background image: Mode-based approach,'' \emph{Transportation research
  record}, vol. 1944, no.~1, pp. 82--88, 2006.

\bibitem{zivkovic2004improved}
Z.~Zivkovic, ``Improved adaptive gaussian mixture model for background
  subtraction,'' in \emph{Proceedings of the 17th International Conference on
  Pattern Recognition, 2004. ICPR 2004.}, vol.~2.\hskip 1em plus 0.5em minus
  0.4em\relax IEEE, 2004, pp. 28--31.

\bibitem{heikkila2006texture}
M.~Heikkila and M.~Pietikainen, ``A texture-based method for modeling the
  background and detecting moving objects,'' \emph{IEEE transactions on pattern
  analysis and machine intelligence}, vol.~28, no.~4, pp. 657--662, 2006.

\bibitem{candes2011robust}
E.~J. Cand{\`e}s, X.~Li, Y.~Ma, and J.~Wright, ``Robust principal component
  analysis?'' \emph{Journal of the ACM (JACM)}, vol.~58, no.~3, pp. 1--37,
  2011.

\bibitem{zhou2012moving}
X.~Zhou, C.~Yang, and W.~Yu, ``Moving object detection by detecting contiguous
  outliers in the low-rank representation,'' \emph{IEEE transactions on pattern
  analysis and machine intelligence}, vol.~35, no.~3, pp. 597--610, 2012.

\bibitem{feng2013online}
J.~Feng, H.~Xu, and S.~Yan, ``Online robust pca via stochastic optimization,''
  in \emph{Advances in Neural Information Processing Systems}, 2013, pp.
  404--412.

\bibitem{shakeri2016corola}
M.~Shakeri and H.~Zhang, ``Corola: A sequential solution to moving object
  detection using low-rank approximation,'' \emph{Computer Vision and Image
  Understanding}, vol. 146, pp. 27--39, 2016.

\bibitem{vaswani2018robust}
N.~Vaswani, T.~Bouwmans, S.~Javed, and P.~Narayanamurthy, ``Robust subspace
  learning: Robust pca, robust subspace tracking, and robust subspace
  recovery,'' \emph{IEEE signal processing magazine}, vol.~35, no.~4, pp.
  32--55, 2018.

\bibitem{bouwmans2018applications}
T.~Bouwmans, S.~Javed, H.~Zhang, Z.~Lin, and R.~Otazo, ``On the applications of
  robust pca in image and video processing,'' \emph{Proceedings of the IEEE},
  vol. 106, no.~8, pp. 1427--1457, 2018.

\bibitem{javed2018moving}
S.~Javed, A.~Mahmood, S.~Al-Maadeed, T.~Bouwmans, and S.~K. Jung, ``Moving
  object detection in complex scene using spatiotemporal structured-sparse
  rpca,'' \emph{IEEE Transactions on Image Processing}, vol.~28, no.~2, pp.
  1007--1022, 2018.

\bibitem{shi2018robust}
G.~Shi, T.~Huang, W.~Dong, J.~Wu, and X.~Xie, ``Robust foreground estimation
  via structured gaussian scale mixture modeling,'' \emph{IEEE Transactions on
  Image Processing}, vol.~27, no.~10, pp. 4810--4824, 2018.

\bibitem{yong2017robust}
H.~Yong, D.~Meng, W.~Zuo, and L.~Zhang, ``Robust online matrix factorization
  for dynamic background subtraction,'' \emph{IEEE transactions on pattern
  analysis and machine intelligence}, vol.~40, no.~7, pp. 1726--1740, 2017.

\bibitem{ebadi2017foreground}
S.~E. Ebadi and E.~Izquierdo, ``Foreground segmentation with tree-structured
  sparse rpca,'' \emph{IEEE transactions on pattern analysis and machine
  intelligence}, vol.~40, no.~9, pp. 2273--2280, 2017.

\bibitem{guo2017godec+}
K.~Guo, L.~Liu, X.~Xu, D.~Xu, and D.~Tao, ``Godec+: fast and robust low-rank
  matrix decomposition based on maximum correntropy,'' \emph{IEEE transactions
  on neural networks and learning systems}, vol.~29, no.~6, pp. 2323--2336,
  2017.

\bibitem{he2011robust}
R.~He, B.-G. Hu, W.-S. Zheng, and X.-W. Kong, ``Robust principal component
  analysis based on maximum correntropy criterion,'' \emph{IEEE Transactions on
  Image Processing}, vol.~20, no.~6, pp. 1485--1494, 2011.

\bibitem{zhan2016online}
J.~Zhan, B.~Lois, H.~Guo, and N.~Vaswani, ``Online (and offline) robust pca:
  Novel algorithms and performance guarantees,'' in \emph{Artificial
  intelligence and statistics}, 2016, pp. 1488--1496.

\bibitem{lois2015online}
B.~Lois and N.~Vaswani, ``Online matrix completion and online robust pca,'' in
  \emph{2015 IEEE International Symposium on Information Theory (ISIT)}.\hskip
  1em plus 0.5em minus 0.4em\relax IEEE, 2015, pp. 1826--1830.

\bibitem{garrigues2010group}
P.~Garrigues and B.~A. Olshausen, ``Group sparse coding with a laplacian scale
  mixture prior,'' in \emph{Advances in neural information processing systems},
  2010, pp. 676--684.

\bibitem{huang2017mixed}
T.~Huang, W.~Dong, X.~Xie, G.~Shi, and X.~Bai, ``Mixed noise removal via
  laplacian scale mixture modeling and nonlocal low-rank approximation,''
  \emph{IEEE Transactions on Image Processing}, vol.~26, no.~7, pp. 3171--3186,
  2017.

\bibitem{dong2018robust}
W.~Dong, T.~Huang, G.~Shi, Y.~Ma, and X.~Li, ``Robust tensor approximation with
  laplacian scale mixture modeling for multiframe image and video denoising,''
  \emph{IEEE Journal of Selected Topics in Signal Processing}, vol.~12, no.~6,
  pp. 1435--1448, 2018.

\bibitem{gep1973bayesian}
B.~Gep and G.~Tiao, ``Bayesian inference in statistical analysis,''
  \emph{Reading: Addison-Wesley}, 1973.

\bibitem{liu2007correntropy}
W.~Liu, P.~P. Pokharel, and J.~C. Pr{\'\i}ncipe, ``Correntropy: Properties and
  applications in non-gaussian signal processing,'' \emph{IEEE Transactions on
  Signal Processing}, vol.~55, no.~11, pp. 5286--5298, 2007.

\bibitem{he2019robust}
Y.~He, F.~Wang, Y.~Li, J.~Qin, and B.~Chen, ``Robust matrix completion via
  maximum correntropy criterion and half-quadratic optimization,'' \emph{IEEE
  Transactions on Signal Processing}, vol.~68, pp. 181--195, 2019.

\bibitem{he2010maximum}
R.~He, W.-S. Zheng, and B.-G. Hu, ``Maximum correntropy criterion for robust
  face recognition,'' \emph{IEEE Transactions on Pattern Analysis and Machine
  Intelligence}, vol.~33, no.~8, pp. 1561--1576, 2010.

\bibitem{du2017robust}
B.~Du, Z.~Wang, L.~Zhang, L.~Zhang, and D.~Tao, ``Robust and discriminative
  labeling for multi-label active learning based on maximum correntropy
  criterion,'' \emph{IEEE Transactions on Image Processing}, vol.~26, no.~4,
  pp. 1694--1707, 2017.

\bibitem{lu2013correntropy}
C.~Lu, J.~Tang, M.~Lin, L.~Lin, S.~Yan, and Z.~Lin, ``Correntropy induced l2
  graph for robust subspace clustering,'' in \emph{Proceedings of the IEEE
  international conference on computer vision}, 2013, pp. 1801--1808.

\bibitem{li2004statistical}
L.~Li, W.~Huang, I.~Y.-H. Gu, and Q.~Tian, ``Statistical modeling of complex
  backgrounds for foreground object detection,'' \emph{IEEE Transactions on
  Image Processing}, vol.~13, no.~11, pp. 1459--1472, 2004.

\bibitem{toyama1999wallflower}
K.~Toyama, J.~Krumm, B.~Brumitt, and B.~Meyers, ``Wallflower: Principles and
  practice of background maintenance,'' in \emph{Proceedings of the seventh
  IEEE international conference on computer vision}, vol.~1.\hskip 1em plus
  0.5em minus 0.4em\relax IEEE, 1999, pp. 255--261.

\bibitem{wang2014cdnet}
Y.~Wang, P.-M. Jodoin, F.~Porikli, J.~Konrad, Y.~Benezeth, and P.~Ishwar,
  ``Cdnet 2014: An expanded change detection benchmark dataset,'' in
  \emph{Proceedings of the IEEE conference on computer vision and pattern
  recognition workshops}, 2014, pp. 387--394.

\bibitem{nikolova2005analysis}
M.~Nikolova and M.~K. Ng, ``Analysis of half-quadratic minimization methods for
  signal and image recovery,'' \emph{SIAM Journal on Scientific computing},
  vol.~27, no.~3, pp. 937--966, 2005.

\bibitem{bertsekas1999nonlinear}
D.~P. Bertsekas, ``Nonlinear programming. athena scientific,'' \emph{Belmont,
  MA}, 1999.

\bibitem{zhou2012bilateral}
T.~Zhou and D.~Tao, ``Bilateral random projections,'' in \emph{2012 IEEE
  International Symposium on Information Theory Proceedings}.\hskip 1em plus
  0.5em minus 0.4em\relax IEEE, 2012, pp. 1286--1290.

\bibitem{he2012incremental}
J.~He, L.~Balzano, and A.~Szlam, ``Incremental gradient on the grassmannian for
  online foreground and background separation in subsampled video,'' in
  \emph{2012 IEEE Conference on Computer Vision and Pattern Recognition}.\hskip
  1em plus 0.5em minus 0.4em\relax IEEE, 2012, pp. 1568--1575.

\bibitem{rodriguez2016incremental}
P.~Rodriguez and B.~Wohlberg, ``Incremental principal component pursuit for
  video background modeling,'' \emph{Journal of Mathematical Imaging and
  Vision}, vol.~55, no.~1, pp. 1--18, 2016.

\bibitem{kang2015robust}
Z.~Kang, C.~Peng, and Q.~Cheng, ``Robust pca via nonconvex rank
  approximation,'' in \emph{2015 IEEE International Conference on Data
  Mining}.\hskip 1em plus 0.5em minus 0.4em\relax IEEE, 2015, pp. 211--220.

\bibitem{wang2012probabilistic}
N.~Wang, T.~Yao, J.~Wang, and D.-Y. Yeung, ``A probabilistic approach to robust
  matrix factorization,'' in \emph{European Conference on Computer
  Vision}.\hskip 1em plus 0.5em minus 0.4em\relax Springer, 2012, pp. 126--139.

\bibitem{zhou2011godec}
T.~Zhou and D.~Tao, ``Godec: Randomized low-rank \& sparse matrix decomposition
  in noisy case,'' in \emph{Proceedings of the 28th International Conference on
  Machine Learning, ICML 2011}, 2011.

\bibitem{wang2004image}
Z.~Wang, A.~C. Bovik, H.~R. Sheikh, and E.~P. Simoncelli, ``Image quality
  assessment: from error visibility to structural similarity,'' \emph{IEEE
  transactions on image processing}, vol.~13, no.~4, pp. 600--612, 2004.

\bibitem{st2014subsense}
P.-L. St-Charles, G.-A. Bilodeau, and R.~Bergevin, ``Subsense: A universal
  change detection method with local adaptive sensitivity,'' \emph{IEEE
  Transactions on Image Processing}, vol.~24, no.~1, pp. 359--373, 2014.

\bibitem{st2016universal}
------, ``Universal background subtraction using word consensus models,''
  \emph{IEEE Transactions on Image Processing}, vol.~25, no.~10, pp.
  4768--4781, 2016.

\bibitem{jiang2017wesambe}
S.~Jiang and X.~Lu, ``Wesambe: A weight-sample-based method for background
  subtraction,'' \emph{IEEE Transactions on Circuits and Systems for Video
  Technology}, vol.~28, no.~9, pp. 2105--2115, 2017.

\bibitem{zheng2012practical}
Y.~Zheng, G.~Liu, S.~Sugimoto, S.~Yan, and M.~Okutomi, ``Practical low-rank
  matrix approximation under robust ${L}_1$-norm,'' in \emph{2012 IEEE
  Conference on Computer Vision and Pattern Recognition}.\hskip 1em plus 0.5em
  minus 0.4em\relax IEEE, 2012, pp. 1410--1417.

\bibitem{ye2015foreground}
X.~Ye, J.~Yang, X.~Sun, K.~Li, C.~Hou, and Y.~Wang, ``Foreground--background
  separation from video clips via motion-assisted matrix restoration,''
  \emph{IEEE Transactions on Circuits and Systems for Video Technology},
  vol.~25, no.~11, pp. 1721--1734, 2015.

\bibitem{zhao2014robust}
Q.~Zhao, D.~Meng, Z.~Xu, W.~Zuo, and L.~Zhang, ``Robust principal component
  analysis with complex noise,'' in \emph{International conference on machine
  learning}, 2014, pp. 55--63.

\bibitem{maddalena2012sobs}
L.~Maddalena and A.~Petrosino, ``The sobs algorithm: What are the limits?'' in
  \emph{2012 IEEE Computer Society Conference on Computer Vision and Pattern
  Recognition Workshops}.\hskip 1em plus 0.5em minus 0.4em\relax IEEE, 2012,
  pp. 21--26.

\bibitem{barnich2010vibe}
O.~Barnich and M.~Van~Droogenbroeck, ``Vibe: A universal background subtraction
  algorithm for video sequences,'' \emph{IEEE Transactions on Image
  processing}, vol.~20, no.~6, pp. 1709--1724, 2010.

\bibitem{cao2015total}
X.~Cao, L.~Yang, and X.~Guo, ``Total variation regularized rpca for irregularly
  moving object detection under dynamic background,'' \emph{IEEE transactions
  on cybernetics}, vol.~46, no.~4, pp. 1014--1027, 2015.

\end{thebibliography}

\end{document}